\documentclass[10pt,twocolumn,letterpaper]{article}

\usepackage{iccv}
\usepackage{times}
\usepackage{epsfig}
\usepackage{graphicx}
\usepackage{amsmath}
\usepackage{amssymb}

\usepackage[pagebackref=true,breaklinks=true,letterpaper=true,colorlinks,bookmarks=false]{hyperref}

\iccvfinalcopy 

\ificcvfinal\fi

\newcommand{\putFigW}[3]{%
  \begin{figure}[tb]%
    \centering%
    \includegraphics[width=#3]{fig/#1.pdf}%
    \caption{#2}%
    \label{fig:#1}%
  \end{figure}}

\newcommand{\putFigWW}[3]{%
  \begin{figure*}[]%
    \centering%
    \includegraphics[width=#3]{fig/#1.pdf}%
    \caption{#2}%
    \label{fig:#1}%
  \end{figure*}}
  
\newcommand{\putFigWWH}[3]{%
  \begin{figure*}[]%
    \centering%
    \includegraphics[width=#3]{fig/#1.pdf}%
    \caption{#2}%
    \label{fig:#1}%
  \end{figure*}}

\newcommand{\refFig}[1]{{Fig. \ref{fig:#1}}}
\newcommand{\refEq}[1]{Eq. (\ref{eq:#1})}

\begin{document}

\title{Enhancement of Novel View Synthesis Using Omnidirectional Image Completion}

\author{Takayuki Hara\\
The University of Tokyo\\
{\tt\small hara@mi.t.u-tokyo.ac.jp}
\and
Tatsuya Harada\\
The University of Tokyo / RIKEN\\
{\tt\small harada@mi.t.u-tokyo.ac.jp}
}

\maketitle
\ificcvfinal\thispagestyle{empty}\fi

\begin{abstract}
In this study, we present a method for synthesizing novel views from a single 360-degree RGB-D image based on the neural radiance field (NeRF) . Prior studies relied on the neighborhood interpolation capability of multi-layer perceptrons to complete missing regions caused by occlusion and zooming, which leads to artifacts. In the method proposed in this study, the input image is reprojected to 360-degree RGB images at other camera positions, the missing regions of the reprojected images are completed by a 360-degree image completion network that is trained in a self-supervised manner to simulate occlusion and resolution changes with viewpoint changes. The completed images are utilized for training NeRF. Because multiple completed images contain inconsistencies in 3D, we introduce a method to learn the NeRF model using a subset of completed images that cover the target scene with less overlap of completed regions. The selection of such a subset of images can be attributed to the maximum weight independent set problem, which is solved through simulated annealing. Experiments demonstrated that the proposed method can synthesize plausible novel views while preserving the features of the scene for both artificial and real-world data.
\end{abstract}


\section{Introduction}

The synthesis of novel views from a set of captured images has a wide range of application, including AR/VR and immersive 3D photography. Conventionally, structure-from-motion \cite{sfm} and image-based rendering \cite{ibr} have been employed for this task. In recent years, neural networks-based rendering methods have been rapidly developed, and the neural radiance field (NeRF) \cite{nerf} is a promising method for synthesizing photorealistic views. However, NeRF requires tens to hundreds of images with known relative positions and the same shooting conditions to be given as the input, and it is a large and time-consuming process. Accordingly, various efforts have been made to reduce the number of input images \cite{pixel-nerf,ibr-net,grf2021,diet-nerf,sparse_fusion} or ease the shooting conditions \cite{nerf-w,nerf--,2021barf,sc-nerf}.

With this background, we attempted to learn a 3D scene model from a single 360-degree image taken in all directions. Learning NeRF from a single 360-degree image is advantageous; that is, we do not need to align the shooting conditions between images or know the relative positions between images. This is because we use only one image that contains massive omnidirectional information. OmniNeRF \cite{omninerf} is a prior study of this approach; however, it relies only on the neighborhood interpolation ability of the multi-layer perceptron to complete the missing regions caused by occlusion and zooming. This leads to artifacts, and the image quality is significantly reduced when moving away from the camera position of the input image.

In contrast, the technology of completing the missing regions of 2D images (i.e., inpainting or image completion) has been studied for a long time. Recent learning-based approaches such as generative adversary networks (GANs) \cite{GAN}, variational autoencoders (VAEs) \cite{VAE,VAE2}, and diffusion models \cite{diff_model1,diff_model2} have made enabled the generation of semantically high-quality images. In addition, 360-degree image generation has been well-researched \cite{panoSyn_j,360ic_j,360ip,s2ic,omni_dreamer,hara2022sig-ss}, and high-quality image generation is possible over the entire field of view. However, 2D image completions generally do not consider the 3D structure; thus, there is no 3D consistency between images completed at multiple camera positions.

In this study, we attempted to synthesize plausible novel views with 3D consistency from a single 360-degree image by combining NeRF and image completion using 360-degree images. Based on a 360-degree image generation model trained in a self-supervised manner, we present a method for completing missing regions caused by occlusion and resolution changes due to viewpoint changes. To maintain 3D consistency, we introduce a method to learn the NeRF model using a subset of completed images that cover the target scene so that the overlap of the completed regions is smaller. Figure \ref{fig:overview} illustrates the overview of our method.

\putFigWW{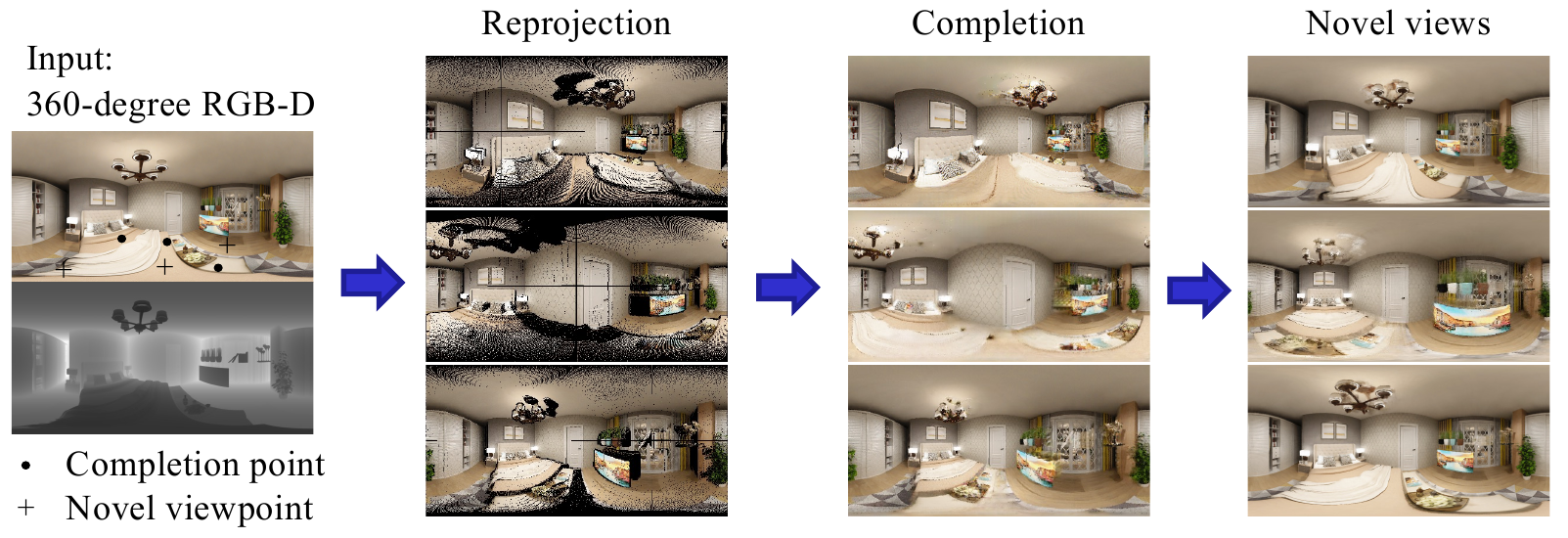}{Given only a single 360 RGB-D image, our method can render novel views. The input image is reprojected to 360-degree images at another camera positions, and the missing regions of the reprojected images are completed. A subset of the selected completed images is used to train the NeRF, and novel views are synthesized. Notably, the 360-degree image is represented by the equirectangular projection that maps the longitude of the viewing direction to the horizontal coordinate and the latitude to the vertical coordinate.}{140mm}

The contributions of this study are as follows.

\begin{itemize}
\item We propose a method for synthesizing novel views by learning NeRF from a single 360-degree RGB-D image, which improves image quality by employing the 360-degree image completion network that is trained in a self-supervised manner to simulate occlusion and resolution changes with viewpoint changes.
\item We designed a new architecture that selects a subset of completed images that cover the target scene with less overlap of completed regions to train the NeRF model, allowing us to maintain 3D consistency for novel views.
\item Our method can be applied to arbitrary NeRF model, and does not require iterations of image completion and NeRF training.
\item We demonstrate that our proposed method can synthesize more plausible views while preserving the features of the scene for both artificial and real-world data.
\end{itemize} 

\putFigWWH{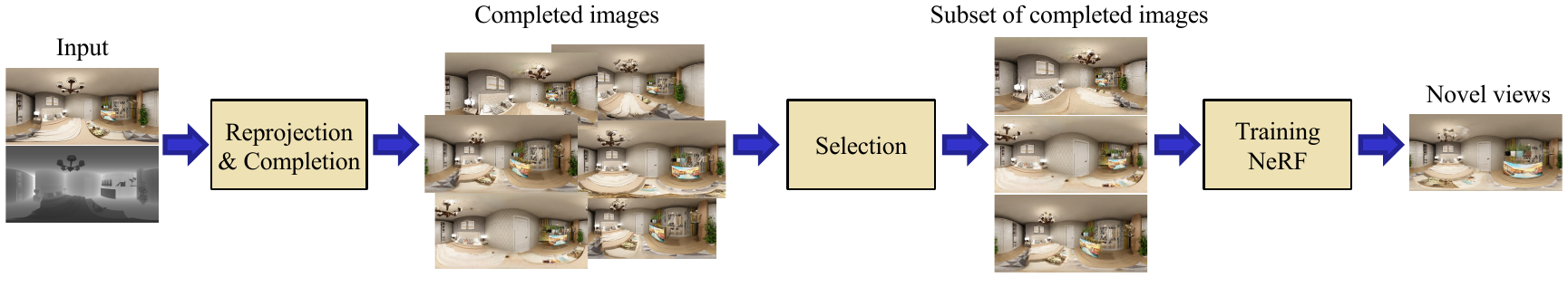}{Pipeline of the proposed method. The input image is reprojected and completed as 360-degree images at other camera positions. A subset of the selected completed images is utilized to train the NeRF. Novel views are then synthesized by the trained NeRF.}{175mm}

\section{Related Work}

\subsection{Novel view synthesis}

Novel view synthesis from a set of captured images has been a consistent challenge in computer vision. Traditionally, structure-from-motion \cite{sfm}, mesh-based methods \cite{kato,meshlet}, multi-plane images (MPI) \cite{mpi,3dp2020,matry}, image-based rendering \cite{dibr,ibr,deep_blending} and light-field photography \cite{light_field} have been studied. There are still problems of image quality limitations and high photographic load.

Recently, rendering techniques using neural networks, which map 3D spatial location to an implicit representation, have been applied to this task \cite{neural_rendering}. In particular, NeRF \cite{nerf} can render objects with complex shapes and textures in a high-quality and photorealistic manner. However, the NeRF requires tens to hundreds of images with known relative positions and the same shooting conditions to be given as input, and such imaging is a large and time-consuming process. Accordingly, various efforts have been made to reduce the number of input images \cite{pixel-nerf,ibr-net,grf2021,diet-nerf,geoaug,sparse_fusion}, ease the shooting conditions \cite{nerf-w,nerf--,2021barf,sc-nerf}, speed up processing \cite{autoint,fast-nerf,kilo-nerf,plenoctrees,hash-nerf,eg3d-gan,mobile-nerf,merf}, and express the entire scene \cite{nerf-pp,2021gsn,mip360,urban-nerf,block-nerf,mega-nerf}.

Among them, OmniNeRF \cite{omninerf} learns an entire scene from a single 360-degree RGB-D image without the need to set relative positions or identify shooting conditions. However, it only relies on the neighborhood interpolation capability of the multi-layer perceptron to complete the missing regions caused by occlusion and zooming, which leads to artifacts. Additionally, the image quality is significantly reduced when moving away from the camera position of the input image. An alternative method to NeRF, pathdreamer \cite{pathdreamer}, synthesizes novel views from a single 360-degree RGB-D image. However, it the method has low 3D consistency in the synthesized views because it is based on 2D image-to-image translation \cite{spade}.

\subsection{Image Completion}

Thus far, various image completion technologies for predicting the missing regions of an image have been proposed. Conventionally, many diffusion-based methods \cite{diffusion1,diffusion2} diffused the information of the visible regions into the missing regions, and multiple patch-based methods \cite{patch1,patch2} completed the missing regions by matching, copying, and realignment using visible regions. Generative models, which are trained using large-scale datasets,  have experienced a significant boost, and the models have been adopted for image completion, with VAE-based methods \cite{PIC,DSI_VQVAE}, GAN-based methods \cite{GAN_IC1,CNN1,CNN2,GAN_IC2,GAN_IC5,GAN_IC6,GAN_IC3,GAN_IC4,UCTGAN,comodgan,pdgan,lama}, autoregressive model-based methods \cite{dii_bat,mat,Dong_2022_CVPR}, and diffusion model-based methods \cite{repaint,ldm,palette}. 

360-degree image generation has also been researched. Han et al. \cite{360ip} proposed an image-inpainting method for spherical structures using a cube map. In \cite{panoSyn,panoSyn_j}, a 360-degree image was generated from a set of images captured in multiple directions as an input. Additionally, panoramic three-dimensional structure prediction methods have been proposed \cite{Im2Pano3D,Lighthouse}. In \cite{IndoorIllum,NeuralIllum,360ic,360ic_j,s2ic,omni_dreamer,hara2022sig-ss}, a 360-degree image was generated from a single normal field of view image. Based on these 360-degree image generation models, we present a self-supervised learning network that completes missing regions caused by occlusion and resolution changes due to changes in viewpoint.

There are also studies of RGB-D image completion\cite{ddcomp,dai2021spsg,pcssc}. RGBD$^2$ \cite{rgbd2} performs repojection and completion incrementally from a few RGB-D images to complete a consistent 3D scene. When this approach is applied to NeRF, the training of NeRF has to be repeated for each image completion, which is computationally very expensive. Therefore, we take the approach of separating NeRF training from image completion, which does not require iterations of NeRF training.

\section{Preliminaries}
\label{sec:pre}

Here, we present an overview of NeRF and OmniNeRF, which forms the basis of this study. NeRF employs a multi-layer perceptron to construct a function that takes the 3D position $x \in \mathbb{R}^3$ and a unit-norm viewing direction $d \in S^2$ as the input, and outputs density $\sigma \in \mathbb{R}$ and color $c \in \mathbb{R}^3$. Using the approximation method of volume rendering, the density and color on the ray that corresponds to the pixel of the image are integrated to calculate the RGB value. Using images captured from multiple viewpoints, the weights of the MLP are learned to minimize the L2 error between the observed and predicted RGB values.

OmniNeRF \cite{omninerf} generates multiple images at virtual camera positions from a single 360-degree RGB-D image and then utilizes these images to train NeRF. A set of 3D points was generated from the given RGB-D panorama and then the 3D points are reprojected into multiple omnidirectional images that correspond to different virtual camera locations. When reprojecting the 3D points onto the virtual camera spheres, their sparsity of the 3D points causes the back part of the object, which is not originally visible to see through. To address this challenge, a median filter was applied to the depth map to mitigate the sparsity.

\section{Proposed Method}

Here, we describe our proposed method that learns the NeRF model and synthesizes novel views from a single 360-degree RGB-D image. Figure \ref{fig:pipeline} illustrates the training pipeline of the proposed method. The input image is first reprojected onto 360-degree RGB images of the virtual camera positions, and the missing regions are completed by a self-supervised 2D image generation model. To maintain 3D consistency, a subset of completed images with less overlap of completed regions is selected. The subset selection is equivalent to the maximum weight independent set problem (MWISP) \cite{mwisp}, which is solved using simulated annealing (SA) \cite{sim_ann}. The input data to train NeRF includes the selected completed images along with RGB data of the input image and its reprojected image. The NeRF is trained to minimize L2 loss of the synthesized and inputed images. 

\subsection{Reprojection and completion}
\label{sec:comp}

\putFigW{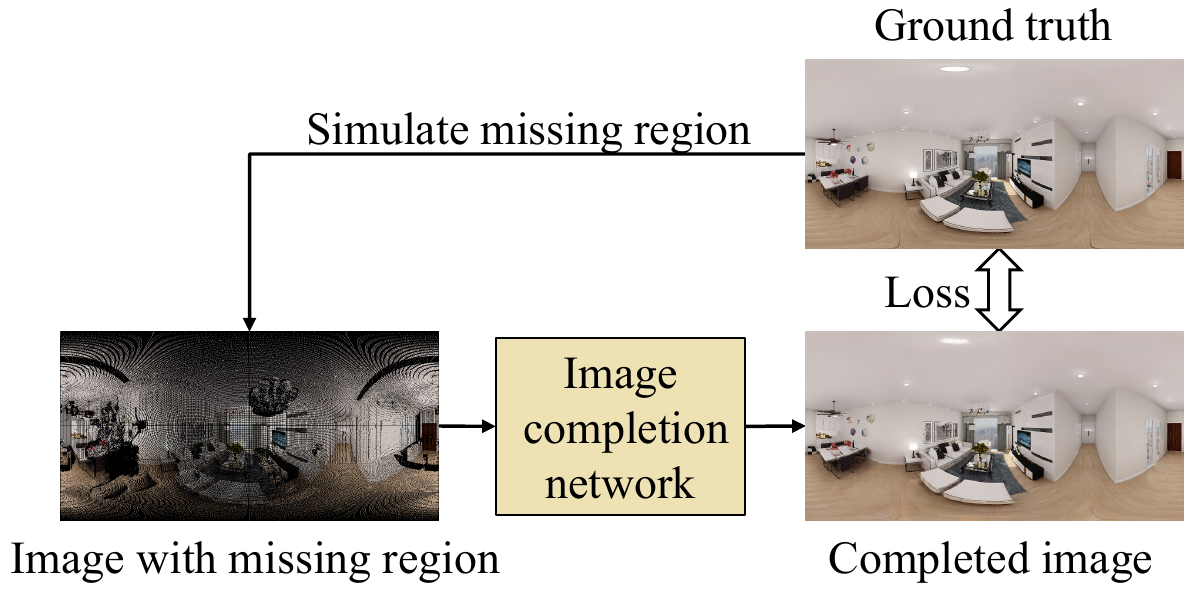}{A mask of the missing region generated by reprojection of the training data is applied to the ground truth image to produce an image that simulates the missing region. The image completion network is trained to minimize the loss between the completed image and the ground truth image.}{80mm}

First, the input image is reprojected onto 360-degree RGB images of the virtual camera positions. This reprojection adopts the same approach as OmniNeRF, as described in Section \ref{sec:pre}. The reprojected image has missing regions owing to occlusion and zooming, as illustrated in \refFig{overview}. We complete the missing regions using an image completion network trained by self-supervised learning manner as shown in \refFig{completion}. Masks of missing regions are generated by reprojecting the 360 RGB-D images of the training data at various positions and applied to the images to train the image completion network. This makes it possible to learn a network on a large number of images with missing regions that may occur, without manual annotations. Our self-supervised learning framework allows the use of any image completion network, and in this paper we employ OmniDreamer \cite{omni_dreamer}, which is a state-of-the-art 360-degree image generation model based on VQ-GAN \cite{vqgan}. The image completion network was trained from scratch for this purpose.

\subsection{Selection of completion positions}

Because image completion is processed in the 360-degree field of view from a single point of view, the consistency within an image is achieved, however the consistency between images observed from different positions cannot be guaranteed. Therefore, a set of images for training the NeRF is adaptively selected from the completed images. It is not easy to determine the combination of the completed images that is 3D consistent. We can determine 3D consistency as a result of training NeRF; however, it takes long to train NeRF once, and it is practically difficult to try all combinations. Therefore, we propose a method for selecting a subset of completed images that cover the target scene; thus, the overlap of the completed regions is smaller in the preliminary stages of training NeRF. 

\subsubsection{Formulation as optimization problem}

 Assuming that there are $N$ completed images, we introduce variable $z_i \in \{0, 1\}$ that takes 1 if $i$-th complete image is selected and 0 if not. Let $r_{ij} \in \mathbb{R}$ be the degree of overlap of the completed regions in completed images $i$ and $j$ (Section \ref{sec:doo}), $R_i$ is a set of rays corresponding to the completed regions in the $i$-th complete image, and $|R_i|$ is the number of elements in $R_i$. We formulate the problem of selecting a subset of completed images for training NeRF as follows:
\begin{eqnarray}
\label{eq:problem}
\mathrm{maximize}_{\{z_i\}_{i=1}^N}&:& \sum_{i=1}^N |R_i| z_i \\
\label{eq:constraint}
\mathrm{subject\ to}&:& r_{ij} < C \ (1 \leq i < j \leq N)
\end{eqnarray}
where $C$ is the threshold parameter that determines overlaps. This optimization problem is equivalent to MWISP. As MWISP is known to be an NP-hard problem \cite{mwisp}, we employ SA \cite{sim_ann} as the optimization method to obtain a suboptimal solution in practical time. The details of the optimization method are described in the Supplemental Material.

\subsubsection{Degree of overlap}
\label{sec:doo}

\putFigW{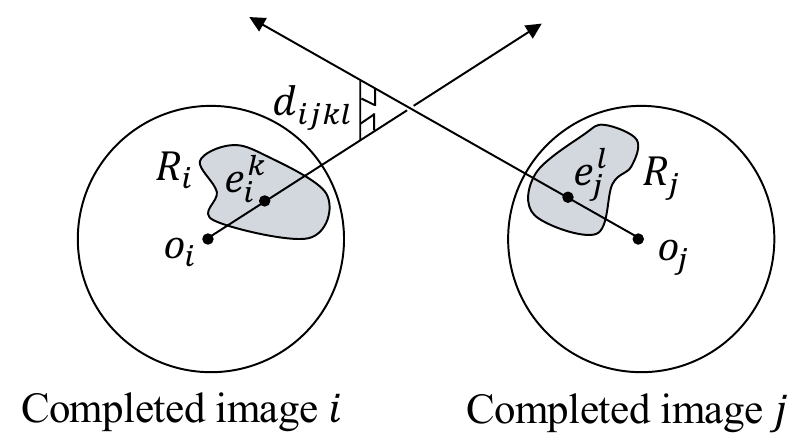}{Determination of collision between two rays.}{60mm}

We formulated the degree of overlap $r_{ij}$ between $i$-th and $j$-th completed images. Determining the overlap between completed regions of different images requires training NeRF, which is time-consuming. Therefore, an assumption is made on the collision probability between rays in the completed region, from which the degree of overlap is estimated.

As shown in \refFig{overlap_explain}, the projection center $i$-th completed image is described as $o_i \in \mathbb{R}^3$, and the direction of the $k$-th ray in $R_i$ is describe as $e_i^k \in S^2$. We assume that the probability that rays $(o_i, e_i^k)$ and $(o_j, e_j^l)$ reflect the same location on the surface is as follows:
\begin{equation}
\label{eq:p_ijkl}
P_{ijkl} = \beta \exp \left( -\frac{d_{ijkl}^2}{2 \sigma^2} \right) \exp \left( \kappa  \langle e_i^k, e_j^l \rangle \right)
\end{equation}
where $\beta$, $\sigma$ and $\kappa$ are hyper-parameters, $\langle \cdot, \cdot \rangle$ is inner product, and $d_{ijkl}$ is the minimum distance between rays $(o_i, e_i^k)$ and $(o_j, e_j^l)$ as follows:
\begin{equation}
\label{eq:d_ijkl}
d_{ijkl} = \frac{ \langle e_i^k \times e_j^l, o_i - o_j \rangle}{||e_i^k \times e_j^l||}.
\end{equation}
Eq. (\ref{eq:p_ijkl}) represents that the co-located reflection probability of the rays is expressed as a Gaussian distribution for the distance between the rays and a von Mises-Fisher distribution for the direction of the rays, which are the standard distributions for distances and angles. Using probability $P_{ijkl}$, the degree of overlap $r_{ij}$ is defined as
\begin{equation}
\label{eq:r_ij}
r_{ij} = \sum_{k \in R_i} \sum_{l \in R_j} P_{ijkl}.
\end{equation}
To speed up the above calculation, a smaller number of pixels are used as $R_i$, randomly sampled from the completed regions.

\subsection{Training NeRF}

The model of NeRF is trained using the selected subset of completed images as the input. Our method can be applied to arbitrary NeRF model since NeRF training and image selection are separated. This has the advantage as NeRF have been improved rapidly in recent years, resulting in a wide variety of methods. In this paper, we employ Instant-NGP \cite{hash-nerf}, which has been used as a baseline in many studies due to its high image quality and fast rendering capability.

\section{Experimental Results}

Quantitative and qualitative experiments were conducted to verify the effectiveness of the proposed method for both synthetic and real-world datasets.

\subsection{Dataset}
\label{sec:dataset}

\subsubsection{Structured3D}
Structured3D dataset \cite{str3d} contains 3,500 synthetic departments (scenes) with 185,985 photorealistic panoramic renderings. As the original virtual environment is not publicly accessible, we utilized the rendered panoramas directly. The data were divided into 3,100 scenes for training, and 400 scenes for testing.

\subsubsection{Matterport3D}
Matterport3D dataset \cite{mat3d} is an indoor real-world 360 dataset, which was captured by Matterport's Pro 3D camera in 90 furnished houses (scenes). The dataset provides 10,800 RGB-D panorama images, and the RGB-D signals near the polar region are missing. The data were divided into 71 scenes for training and 19 scenes for testing.

\subsection{Implementation Details}
\label{sec:impl}

The input images, completed images, and novel views were in equirectangular projection format with a resolution of $1,024 \times 512$. The reprojection and completion positions are taken at 10 $\times$ 10 grid points on the $x$-$y$ horizontal plane by equally dividing the interval $[\alpha x_{\rm min}, \alpha x_{\rm max}]$ on the x-axis and the interval  $[\alpha y_{\rm min}, \alpha y_{\rm max}]$ on the y-axis,  where $\alpha=0.6$ and $x_{\rm min}, x_{\rm max}, y_{\rm min}, y_{\rm max}$ are the boundary point of the depth of the input image. For the selection of completion positions, the hyper-parameters of $P_{ijkl}$ are set as $\beta = 1.0$, $\sigma = 0.01$ and $\kappa=1.0$,  overlap threshold $C$ is set to $4.0 \times 10^4$, and the sampling rate of the rays for $R_i$ is set to 0.01.

The image completion networks were trained on training data, whereas NeRF was trained on test data, according to the division defined in Section \ref{sec:dataset}. The 360-degree RGB-D images were reprojected to generate 3,200 masks of missing regions, which were applied to the images on training data for training the image completion network. We trained the NeRF model with 200,000 iterations for each experiment with a batch size of 1,400. The network structures of NeRF were identical to those of Instant-NGP \cite{hash-nerf}. OmniNeRF \cite{omninerf} was used for comparison, and the NeRF model and training settings are identical to those of the proposed method. The details of the models and training configurations are described in the Supplemental Material. 

\putFigWWH{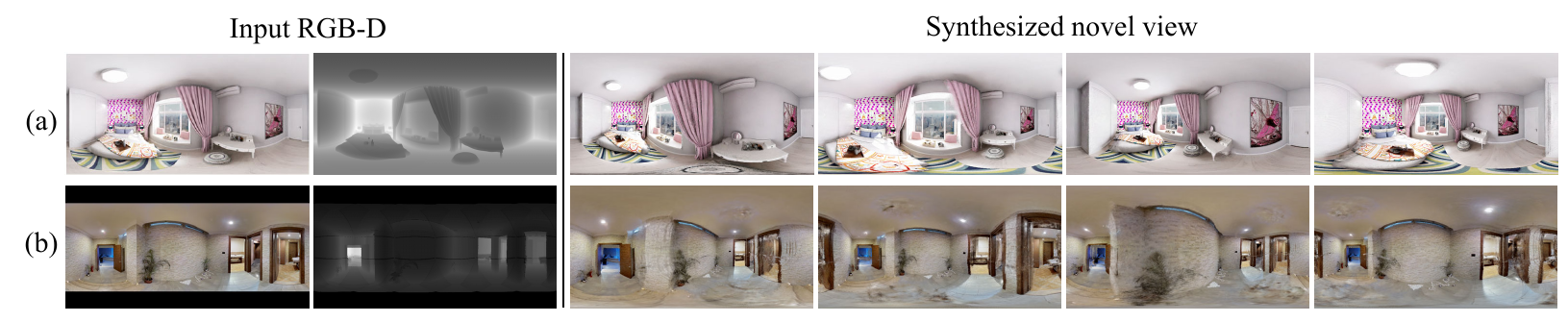}{Visualization of novel view rendering in the proposed method. (a) A sample in the Structure3D dataset. (b) A sample in the Matterport3D dataset. A plausible viewpoint image with 3D consistency is synthesized at a position different from the input image.}{175mm}

\putFigWW{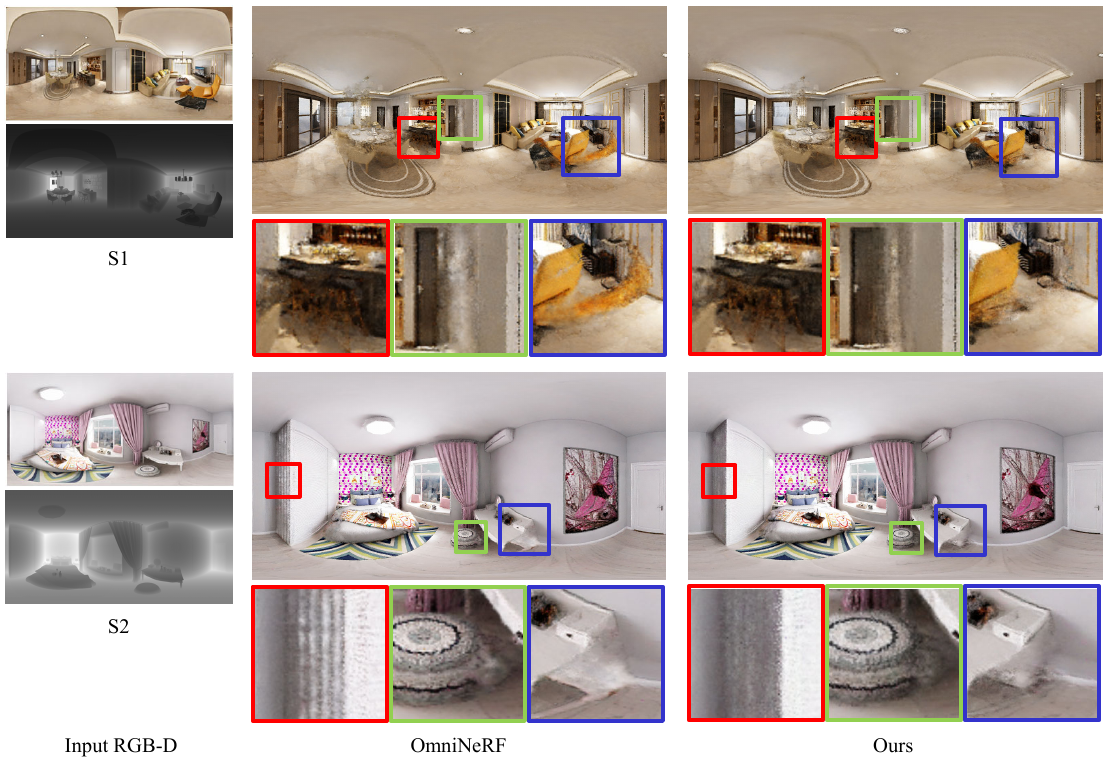}{Qualitative comparison of OmniNeRF and the proposed method on the Structure3D dataset. OmniNeRF produces artifacts in occlusion regions such as behind the yellow chair in the blue bounding box of scene S1 and on the wall in the red bounding box of  scene S2, whereas the proposed method reduces these artifacts. The backless chair in the green bounding box of scene S1 has a collapsed image in OmniNerf owing to changes in the resolution and the viewing angle, whereas the proposed method reduces shape collapse.}{170mm}

\putFigWW{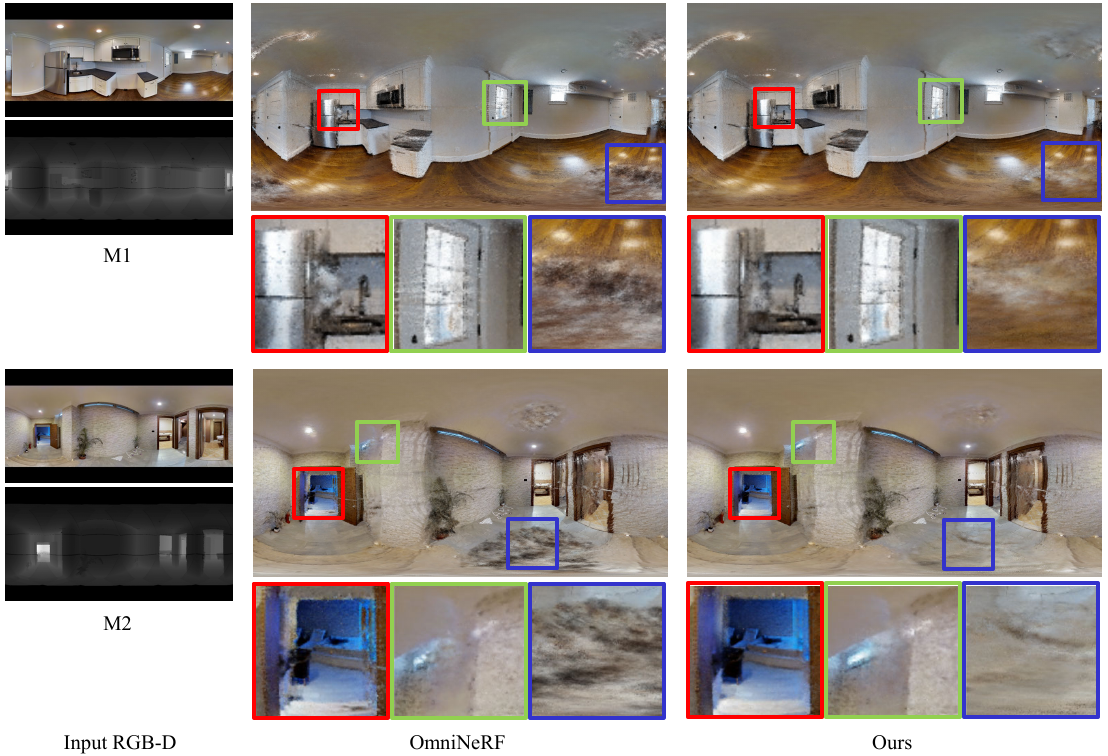}{Qualitative comparison of each novel view synthesis method on the Matterport3D dataset. The Matterport3D dataset contains missing regions at the top and bottom regions of the input image, and the proposed method completes the missing regions naturally, as in the floor of scenes M1 and M2. The refrigerator and the window in scene M1 have some artifacts in OmniNeRF owing to resolution change; however, the proposed method reduces these artifacts.}{170mm}

\putFigWW{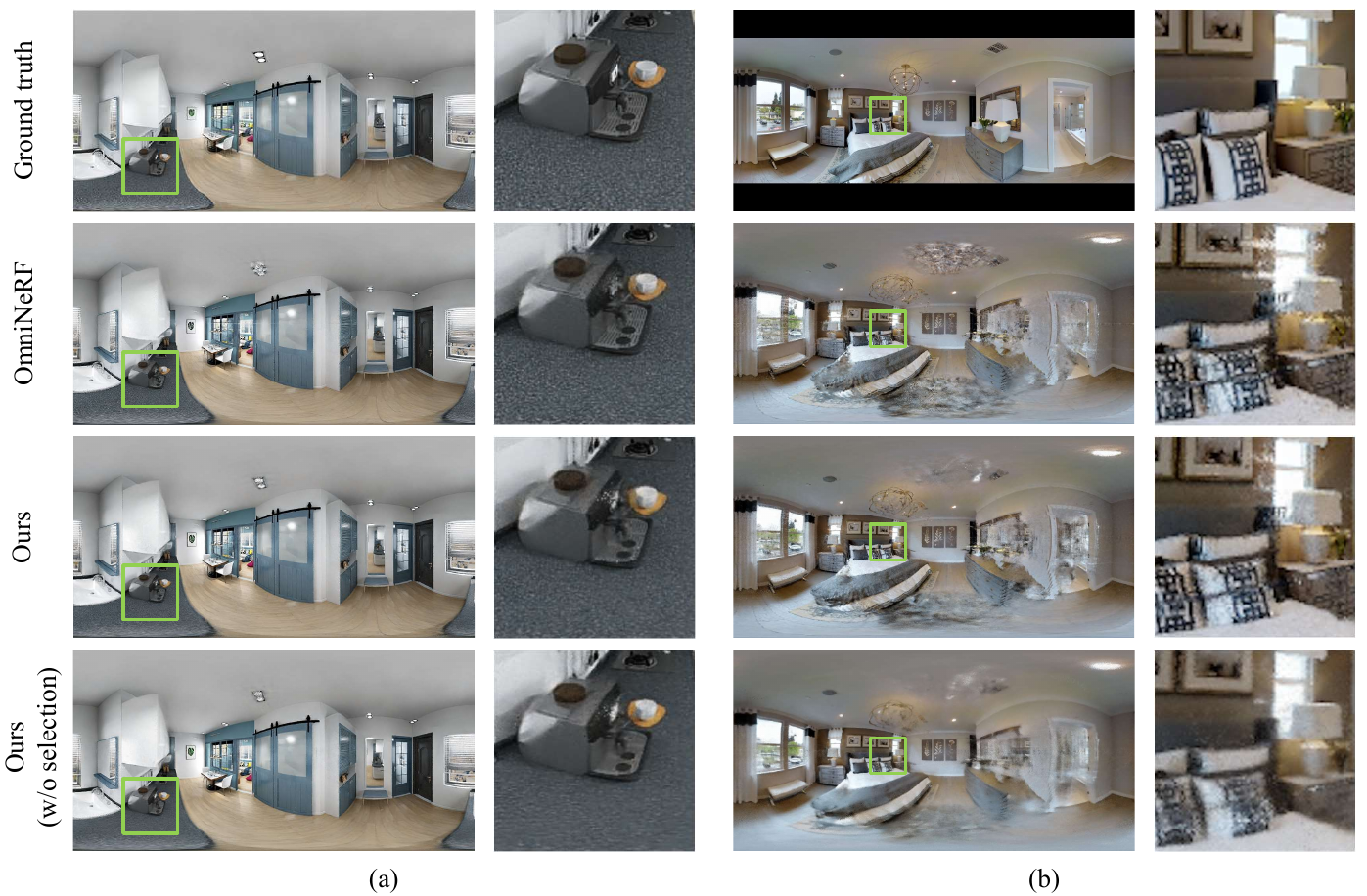}{Comparison with the ground truth images.  (a) A sample in the Structure3D dataset, where the input image is used for the ground truth image. (b) A sample in the Matterport3D dataset,  where the image closest to the input image is used for the ground truth image.}{170mm}

\subsection{Qualiative evaluation}

First, we qualitatively validate the novel view synthesis using a single 360-degrees RGB-D image. Figure \ref{fig:trans} illustrates examples of synthesized novel views using the proposed method. A plausible view with 3D consistency is synthesized at a position different from the camera position of the input. The depth data in the Matterport3D dataset is subject to measurement error, which results in some distortion in the synthesized views. In \refFig{sample_str3d} and \refFig{sample_mat3d}, we compare images synthesized using OmniNeRF and the proposed method. OmniNeRF produces artifacts and noise in the occlusion regions and in the area where the resolution has changed, whereas the proposed method produces a plausible image. These results indicate that the Image completion is working effectively. Additional results are available in the Supplementary Material.

\subsection{Quantitative evaluation}

\begin{table}[tb]
\caption{Evaluation results of novel view synthesis  in Structured3D dataset \cite{str3d}.}
\label{table:quant_s3d}
 \centering
 \scalebox{0.86}{
  \begin{tabular}{p{8.5em} | p{3.2em}   } \hline
 \hfil Method & \hfil NLL$\downarrow$ \\  \hline \hline
 \hfil OmniNeRF & \hfil 2.316 \\ 
 \hfil  Ours & \hfil \underline{2.295} \\
 \hfil  Ours (w/o selection) & \hfil \textbf{2.293} \\ \hline
  \end{tabular}
  }
\end{table}

\begin{table}[tb]
\caption{Evaluation results of novel view synthesis  in Materport3D dataset \cite{mat3d}.}
\label{table:quant_m3d}
 \centering
 \scalebox{0.86}{
  \begin{tabular}{p{8.5em} | p{3.2em}   p{3.2em}  p{3.2em}  p{3.2em} } \hline
 \hfil Method &  \hfil  PSNR$\uparrow$ & \hfil SSIM$\uparrow$ & \hfil LPIPS$\downarrow$ & \hfil NLL$\downarrow$\\  \hline \hline
 \hfil OmniNeRF  & \hfil 14.917 & \hfil 0.394 & \hfil  \underline{0.551} & \hfil \underline{2.055} \\ 
 \hfil  Ours & \hfil \underline{14.976} & \hfil \underline{0.401} & \hfil  \textbf{0.547} & \hfil \textbf{2.024} \\
 \hfil  Ours (w/o selection) & \hfil 1\textbf{5.028} & \hfil \textbf{0.402} & \hfil  0.553 & \hfil 2.113\\  \hline
  \end{tabular}
  }
\end{table}

\subsubsection{Evaluation metrics}

We quantitatively evaluated each method using the following four evaluation metrics: the peak signal-to-noise ratio (PSNR), SSIM \cite{ssim}, LPIPS \cite{lpips} and the negative log likelihood (NLL). PSNR, SSIM, and LPIPS are calculated between the synthesized and the ground truth images. NLL evaluates the plausibility of the synthesized views for the feature distribution of test data. Using the values of the last pooling layer of inception-v3 \cite{inception_v3} as features, the likelihood of the synthesized image is calculated for the feature distribution of the test data. A detailed definition and validity are provided in the Supplementary Material. Although FID score \cite{fid} is standard for evaluating image generation, it requires more than 1000 images to obtain reliable results, making it inappropriate for use in this study, where it takes more than 4 hours to generate a single scene.

\subsubsection{Evaluation results}

To verify the effectiveness of our method in various scenes, we selected 22 and 19 images for the evaluation from the test data of the Structure3D and Matterport3D respectively, and calculated the mean of the evaluation metrics. The evaluation results of novel view synthesis for each dataset are presented in Table \ref{table:quant_s3d} and \ref{table:quant_m3d}. Note that the Structured3D dataset does not have the ground truth image at the novel viewpoint, therefore only the NLL is used for evaluation.The proposed method outperforms OmniNeRF on all datasets and all evaluation metrics. This results reflects the reduction of artifacts and blurring in the regions of occlusion and resolution change, as seen in the qualitative evaluation. This could be because the image completion network trained on the training data generalized image features and it plausibly completed the missing regions in the scene. 

\subsection{Ablation study}

We conduct an ablation study with learning NeRF using all 100 completed images without selection. The results are presented in Table  \ref{table:quant_s3d} and \ref{table:quant_m3d}. Without the image selection, only PSNR for the Matterport3D dataset is clearly better, while LPIPS and NLL for the same dataset are even worse than OmniNeRF. On the other hand, the method with the image selection outperforms OmniNeRF in all evaluation metrics. From this perspective, we can see the usefulness of the image selection.


\refFig{gt_rec} shows a comparison of the ground truth images and the images synthesized by each method. Since the Structured3D dataset does not contain enough overlapping images, we use the input image as the ground truth image for comparison with the synthesized image at the input image position. In the Matterport3D dataset, we use the image taken at the position closest to the input image as the ground truth image. Without the image selection, the synthesized image is prone to blurring. This is due to the reason that the completed images with 3D inconsistencies were used to train the NeRF, resulting in synthesizing average images. Image blurring degrades LPIPS which evaluates semantic similarity and NLL which evaluates image plausibility in the Matterport3D dataset. The NLL in the Structured3D dataset was almost the same with and without image selection. This may be due to the reason that Structured3D is an artificial data set with relatively little texture and is less susceptible to blurring, and that the scene is narrower and the missing areas are on average 16\% smaller than in Matterport3D, resulting in fewer inconsistencies due to image completion. The reason that PSNR and SSIM are equal or better than method with image selection can be attributed to the fact that the image blurring leads to robustness against the positional shift under conditions where the evaluation position is far from the input image position and difficult to reproduce with high positional accuracy.

\subsection{Limitations}

Although the performance of the proposed method is promising, it has several limitations. First, if there are large missing regions that exceed the image completion capabilities, it is difficult to synthesize plausible views. Second, the reprojection process highly depends on the depth accuracy, and geometric distortion occurs in the synthesized image when the depth accuracy is low.

\section{Conclusions}

In this study, we propose a method for synthesizing novel views by learning the NeRF from a single 360-degree RGB-D image. Unlike existing methods \cite{omninerf}, the proposed method employed the 360-degree image completion network that is trained in a self-supervised manner to simulate occlusion and resolution changes with viewpoint changes. The completed images for reprojected images at other camera positions are utilized for training the NeRF. To avoid 3D inconsistencies, we introduced a method to train NeRF using a subset of completed images that cover the target scene with less overlap of completed regions. 

The experiments indicated that the proposed method can synthesize plausible novel views while preserving the features of the scene, both for artificial and real-world data. These results confirm the effectiveness of employing image completion and the selection of a subset of the completed image with consistency for novel view synthesis. Recently, a method for training NeRF from a single 360-degree RGB-D image was proposed, and it used a large vision-language model to maintain consistency of novel viewpoints \cite{360fusion-nerf}. Our method has the advantage of being lightweight in processing, and it can be combined with such methods to improve quality since the proposed method can employ arbitrary NeRF model.

\section*{Acknowledgements}

This work was partially supported by JST AIP Acceleration Research JPMJCR20U3, Moonshot R\&D Grant Number JPMJPS2011, CREST Grant Number JPMJCR2015, JSPS KAKENHI Grant Number JP19H01115, JP20H05556 and Basic Research Grant (Super AI) of Institute for AI and Beyond of the University of Tokyo.

{\small
\bibliographystyle{ieee_fullname}
\bibliography{hara}

\begin{thebibliography}{10}\itemsep=-1pt

\bibitem{360ic_j}
Naofumi Akimoto and Yoshimitsu Aoki.
\newblock Image completion of 360-degree images by cgan with residual
  multi-scale dilated convolution.
\newblock {\em IIEEJ Trans. Image Electronics and Vis. Comput.}, 8(1):35--43,
  2020.

\bibitem{omni_dreamer}
Naofumi Akimoto, Yuhi Matsuo, and Yoshimitsu Aoki.
\newblock Diverse plausible 360-degree image outpainting for efficient 3dcg
  background creation.
\newblock In {\em Proc. IEEE Conf. Comput. Vis. Pattern Recognit.}, pages
  11441--11450, 2022.

\bibitem{matry}
Benjamin Attal, Selena Ling, Aaron Gokaslan, Christian Richardt, and James
  Tompkin.
\newblock Matryodshka: Real-time 6dof video view synthesis using multi-sphere
  images.
\newblock In {\em Europ. Conf. Comput. Vis.}, 2020.

\bibitem{meshlet}
Abhishek Badki, Orazio Gallo, Jan Kautz, and Pradeep Sen.
\newblock Meshlet priors for 3d mesh reconstruction.
\newblock In {\em Proc. IEEE Conf. Comput. Vis. Pattern Recognit.}, page
  2846–2855, 2020.

\bibitem{diffusion1}
Coloma Ballester, Marcelo Bertalmio, Vicent Caselles, Guillermo Sapiro, and
  Joan Verdera.
\newblock Filling-in by joint interpolation of vector fields and gray levels.
\newblock {\em IEEE Trans. Image Processing}, 10(8):1200--1211, 2001.

\bibitem{patch2}
Connelly Barnes, Eli Shechtman, Adam Finkelstein, and Dan~B Goldman.
\newblock Patchmatch: A randomized correspondence algorithm for structural
  image editing.
\newblock {\em ACM Trans. Graph.}, 28(3), 2009.

\bibitem{mip360}
Jonathan~T. Barron, Ben Mildenhall, Dor Verbin, Pratul~P. Srinivasan, and Peter
  Hedman.
\newblock Mip-nerf 360: Unbounded anti-aliased neural radiance fields.
\newblock {\em arXiv:2111.12077}, 2021.

\bibitem{diffusion2}
Marcelo Bertalmio, Guillermo Sapiro, Vincent Caselles, and Coloma Ballester.
\newblock Image inpainting.
\newblock In {\em Proc. Conf. Comput. Graph. Interact. Techniq.}, pages
  417--424, 2000.

\bibitem{eg3d-gan}
Eric~R. Chan, Connor~Z. Lin, Matthew~A. Chan, Koki Nagano, Boxiao Pan,
  Shalini~De Mello, Orazio Gallo, Leonidas Guibas, Jonathan Tremblay, Sameh
  Khamis, Tero Karras, and Gordon Wetzstein.
\newblock Efficient geometry-aware 3d generative adversarial networks.
\newblock {\em arXiv:2112.07945}, 2020.

\bibitem{mat3d}
Angel Chang, Angela Dai, Thomas Funkhouser, Maciej Halber, Matthias Nießner,
  Manolis Savva, Shuran Song, Andy Zeng, and Yinda Zhang.
\newblock Matterport3d: Learning from rgb-d data in indoor environments.
\newblock In {\em Int. Confe. 3D Vis.}, 2017.

\bibitem{geoaug}
Di Chen, Yu Liu, Lianghua Huang, Bin Wang, and Pan Pan.
\newblock Geoaug: Data augmentation for few-shot nerf with geometry
  constraints.
\newblock In {\em Europ. Conf. Comput. Vis.}, 2022.

\bibitem{mobile-nerf}
Zhiqin Chen, Thomas Funkhouser, Peter Hedman, and Andrea Tagliasacchi.
\newblock Mobilenerf: Exploiting the polygon rasterization pipeline for
  efficient neural field rendering on mobile architectures.
\newblock {\em arXiv:2208.00277}, 2022.

\bibitem{patch1}
Antonio Criminisi, Patrick Perez, and Kentaro Toyama.
\newblock Object removal by exemplar-based inpainting.
\newblock In {\em Proc. IEEE Conf. Comput. Vis. Pattern Recognit.}, pages
  II--II, 2003.

\bibitem{dai2021spsg}
Angela Dai, Yawar Siddiqui, Justus Thies, Julien Valentin, and Matthias
  Nie{\ss}ner.
\newblock Spsg: Self-supervised photometric scene generation from rgb-d scans.
\newblock In {\em Proc. IEEE Conf. Comput. Vis. Pattern Recognit.}, 2021.

\bibitem{2021gsn}
Terrance DeVries, Miguel~Angel Bautista, Nitish Srivastava, Graham~W. Taylor,
  and Joshua~M. Susskind.
\newblock Unconstrained scene generation with locally conditioned radiance
  fields.
\newblock In {\em Proc. IEEE Int. Conf. Comput. Vis.}, 2021.

\bibitem{Dong_2022_CVPR}
Qiaole Dong, Chenjie Cao, and Yanwei Fu.
\newblock Incremental transformer structure enhanced image inpainting with
  masking positional encoding.
\newblock In {\em Proc. IEEE Conf. Comput. Vis. Pattern Recognit.}, 2022.

\bibitem{vqgan}
Patrick Esser, Robin Rombach, and Bjorn Ommer.
\newblock Taming transformers for high-resolution image synthesis.
\newblock In {\em Proc. IEEE Conf. Comput. Vis. Pattern Recognit.}, 2021.

\bibitem{dibr}
Christoph Fehn.
\newblock Depth-image-based rendering (dibr), compression, and transmission for
  a new approach on 3d-tv.
\newblock In {\em Proc. Stereoscopic Displays and Virtual Reality Systems XI},
  2004.

\bibitem{CNN1}
K. Fukushima and S Miyake.
\newblock Neocognitron: A new algorithm for pattern recognition tolerant of
  deformations and shifts in position.
\newblock {\em Pattern Recognition}, 15(6):455--469, 1982.

\bibitem{fast-nerf}
Stephan~J. Garbin, Marek Kowalski, Matthew Johnson, Jamie Shotton, and Julien
  Valentin.
\newblock Fastnerf: High-fidelity neural rendering at 200fps.
\newblock In {\em International Conference on Computer Vision}, 2021.

\bibitem{IndoorIllum}
Marc-Andre Gardner, Kalyan Sunkavalli, Ersin Yumer, Xiaohui Shen, Emiliano
  Gambaretto, Gagne Christian, and Jean-Francois Lalonde.
\newblock Learning to predict indoor illumination from a single image.
\newblock {\em ACM Trans. Graph.}, 9(4), 2017.

\bibitem{GAN}
Ian Goodfellow, Jean Pouget-Abadie, Mehdi Mirza, Bing Xu, David Warde-Farley,
  Sherjil Ozair, Aaronand Courville, and Yoshua Bengio.
\newblock Generative adversarial nets.
\newblock In {\em Proc. Int. Conf. Neural Inf. Process. Syst.}, pages
  2672--2680., 2014.

\bibitem{360ip}
S.~W. Han and D.~Y. Suh.
\newblock A 360-degree panoramic image inpainting network using a cube map.
\newblock {\em CMC-Computers, Materials and Continua}, 66(1):213--228, 2021.

\bibitem{s2ic}
Takayuki Hara, Yusuke Mukuta, and Tatsuya Harada.
\newblock Spherical image generation from a single image by considering scene
  symmetry.
\newblock In {\em Proc. AAAI Conf. Artif. Intell.}, pages 1513--1521, 2021.

\bibitem{hara2022sig-ss}
Takayuki Hara, Yusuke Mukuta, and Tatsuya Harada.
\newblock Spherical image generation from a few normal-field-of-view images by
  considering scene symmetry.
\newblock {\em IEEE Transactions on Pattern Analysis and Machine Intelligence},
  pages 1--15, 2022.

\bibitem{sfm}
Richard Hartley and Andrew Zisserman.
\newblock {\em Multiple view geometry in computer vision}.
\newblock Cambridge university press, 2003.

\bibitem{deep_blending}
Peter Hedman, Julien Philip, True Price, Jan-Michael Frahm, George Drettakis,
  and Gabriel Brostow.
\newblock Deep blending for free-viewpoint image-based rendering.
\newblock 37(6):1--15, 2018.

\bibitem{fid}
Martin Heusel, Hubert Ramsauer, Thomas Unterthiner, and Bernhard Nessler.
\newblock Gans trained by a two time-scale update rule converge to a local nash
  equilibrium.
\newblock In {\em Proc. Int. Conf. Neural Inf. Process. Syst.}, page
  6626–6637, 2017.

\bibitem{diff_model2}
Jonathan Ho, Ajay Jain, and Pieter Abbeel.
\newblock Denoising diffusion probabilistic models.
\newblock In {\em Proc. Int. Conf. Neural Inf. Process. Syst.}, 2020.

\bibitem{omninerf}
Ching-Yu Hsu, Cheng Sun, and Hwann-Tzong Chen.
\newblock Moving in a 360 world: Synthesizing panoramic parallaxes from a
  single panorama.
\newblock {\em arXiv:2106.10859}, 2021.

\bibitem{GAN_IC2}
SATOSHI Iizuka, Edgar Simo-Serra, and HIROSHI Ishikawa.
\newblock Globally and locally consistent image completion.
\newblock {\em ACM Trans. Graph.}, 36(4), 2017.

\bibitem{diet-nerf}
Ajay Jain, Matthew Tancik, and Pieter Abbeel.
\newblock Putting nerf on a diet: Semantically consistent few-shot view
  synthesiss.
\newblock In {\em Proc. IEEE Int. Conf. Comput. Vis.}, 2021.

\bibitem{sc-nerf}
Yoonwoo Jeong, Seokjun Ahn, Christopher Choy, Animashree Anandkumar, Minsu Cho,
  and Jaesik Park.
\newblock Self-calibrating neural radiance fields.
\newblock In {\em International Conference on Computer Vision}, 2021.

\bibitem{kato}
Hiroharu Kato, Yoshitaka Ushiku, and Tatsuya Harada.
\newblock Neural 3d mesh renderer.
\newblock In {\em Proc. IEEE Conf. Comput. Vis. Pattern Recognit.}, page
  3907–3916, 2018.

\bibitem{adam}
Diederik~P. Kingma and Jimmy~Lei Ba.
\newblock Adam: A method for stochastic optimization.
\newblock {\em arXiv:1412.6980}, 2014.

\bibitem{VAE}
Diederik~P Kingma and Max Welling.
\newblock Auto-encoding variational bayes.
\newblock {\em arXiv:1312.6114}, 2013.

\bibitem{sim_ann}
Scott Kirkpatrick, C.~D. Gelatt~Jr, and M.~P. Vecchi.
\newblock Optimization by simulated annealing.
\newblock {\em Science}, 220(4598):671--680, 1983.

\bibitem{pathdreamer}
Jing~Yu Koh, Honglak Lee, Yinfei Yang, Jason Baldridge, and Peter Anderson.
\newblock Pathdreamer: A world model for indoor navigation.
\newblock In {\em Proc. IEEE Int. Conf. Comput. Vis.}, 2021.

\bibitem{360fusion-nerf}
Shreyas Kulkarni, Peng Yin, and Sebastian Scherer.
\newblock 360fusionnerf: Panoramic neural radiance fields with joint guidance.
\newblock {\em arXiv:2209.14265}, 2022.

\bibitem{CNN2}
Y. LeCun, B. Boser, J.~S. Denker, D. Henderson, R.~E. Howard, W. Hubbard, and
  L.~D. Jackel.
\newblock Backpropagation applied to handwritten zip code recognition.
\newblock {\em Neural computation}, 1(4):541--551, 1989.

\bibitem{rgbd2}
Jiabao Lei, Jiapeng Tang, and Kui Jia.
\newblock Rgbd2: Generative scene synthesis via incremental view inpainting
  using rgbd diffusion models.
\newblock In {\em Proc. IEEE Conf. Comput. Vis. Pattern Recognit.}, 2023.

\bibitem{light_field}
Marc Levoy and Pat Hanrahan.
\newblock Light field rendering.
\newblock In {\em SIGGRAPH}, 2020.

\bibitem{mat}
Wenbo Li, Zhe Lin, Kun Zhou, Lu Qi, Yi Wang, and Jiaya Jia.
\newblock Mat: Mask-aware transformer for large hole image inpainting.
\newblock In {\em Proc. IEEE Conf. Comput. Vis. Pattern Recognit.}, 2022.

\bibitem{GAN_IC1}
Yijun Li, Sifei Liu, Jimei Yang, and Ming-Hsuan Yang.
\newblock Generative face completion.
\newblock In {\em Proc. IEEE Conf. Comput. Vis. Pattern Recognit.}, pages
  3911--3919, 2017.

\bibitem{2021barf}
Chen-Hsuan Lin, Wei-Chiu Ma, Antonio Torralba, and Simon Lucey.
\newblock Barf : Bundle-adjusting neural radiance fields.
\newblock In {\em International Conference on Computer Vision}, 2021.

\bibitem{GAN_IC5}
G. Liu, F.~A. Reda, K.~J. Shih, T.~C. Wang, A. Tao, and B. Catanzaro.
\newblock Image inpainting for irregular holes using partial convolutions.
\newblock In {\em Proc. Eur. Conf. Comput. Vis.}, pages 85--100, 2018.

\bibitem{pdgan}
H. Liu, Z. Wan, W. Huang, Y. Song, X. Han, and J. Liao.
\newblock Pd-gan: Probabilistic diverse gan for image inpainting.
\newblock In {\em Proc. IEEE Conf. Comput. Vis. Pattern Recognit.}, pages
  9371--9381, 2021.

\bibitem{repaint}
Andreas Lugmayr, Martin Danelljan, Andres Romero, Fisher Yu, Radu Timofte, and
  Luc~Van Gool.
\newblock Repaint: Inpainting using denoising diffusion probabilistic models.
\newblock 2022.

\bibitem{nerf-w}
Ricardo Martin-Brualla, Noha Radwan, Mehdi S.~M. Sajjadi, Jonathan~T. Barron,
  Alexey Dosovitskiy, and Daniel Duckworth.
\newblock Nerf in the wild: Neural radiance fields for unconstrained photo
  collections.
\newblock In {\em Proc. IEEE Conf. Comput. Vis. Pattern Recognit.}, 2021.

\bibitem{nerf}
Ben Mildenhall, Pratul~P. Srinivasan, Matthew Tancik, Jonathan~T. Barron, Ravi
  Ramamoorthi, and Ren Ng.
\newblock Nerf: Representing scenes as neural radiance fields for view
  synthesis.
\newblock In {\em Europ. Conf. Comput. Vis.}, 2020.

\bibitem{hash-nerf}
Thomas M\"uller, Alex Evans, Christoph Schied, and Alexander Keller.
\newblock Instant neural graphics primitives with a multiresolution hash
  encoding.
\newblock {\em arXiv:2201.05989}, Jan. 2022.

\bibitem{360ic}
Akimoto Naofumi, Seito Kasai, Masaki Hayashi, and Yoshimitsu Aoki.
\newblock 360-degree image completion by two-stage conditionalgans.
\newblock In {\em Proc. IEEE Int. Conf. Image Processing}, pages 4704--4708,
  2019.

\bibitem{spade}
Taesung Park, Ming-Yu Liu, Ting-Chun Wang, and Jun-Yan Zhu.
\newblock Semantic image synthesis with spatially-adaptive normalization.
\newblock In {\em Proc. IEEE Conf. Comput. Vis. Pattern Recognit.}, 2019.

\bibitem{DSI_VQVAE}
Jialun Peng, Dong Liu, Songcen Xu, and Houqiang Li.
\newblock Generating diverse structure for image inpainting with hierarchical
  vq-vae.
\newblock In {\em Proc. IEEE Conf. Comput. Vis. Pattern Recognit.}, pages
  10775--10784, 2021.

\bibitem{mwisp}
Wayne Pullan.
\newblock Optimisation of unweighted/weighted maximum independent sets and
  minimum vertex covers.
\newblock {\em Discrete Optimization}, 6(2):214--219, 2009.

\bibitem{kilo-nerf}
Christian Reiser, Songyou Peng, Yiyi Liao, and Andreas Geiger.
\newblock Kilonerf: Speeding up neural radiance fields with thousands of tiny
  mlps.
\newblock In {\em Proc. IEEE Int. Conf. Comput. Vis.}, Jan. 2021.

\bibitem{merf}
Christian Reiser, Richard Szeliski, Dor Verbin, Pratul~P. Srinivasan, Ben
  Mildenhall, Andreas Geiger, Jonathan~T. Barron, and Peter Hedman.
\newblock Merf: Memory-efficient radiance fields for real-time view synthesis
  in unbounded scenes.
\newblock {\em arXiv:2208.00277}, 2023.

\bibitem{urban-nerf}
Konstantinos Rematas, Andrew Liu, Pratul~P. Srinivasan, Jonathan~T. Barron,
  Andrea Tagliasacchi, Thomas Funkhouser, and Vittorio Ferrari.
\newblock Urban radiance fields.
\newblock {\em arXiv:2111.14643}, 2021.

\bibitem{VAE2}
Danilo~Jimenez Rezende, Shakir Mohamed, and Daan Wierstra.
\newblock Stochastic backpropagation and approximate inference in deep
  generative models.
\newblock In {\em Proc. Int. Conf. Mach. Learn.}, pages 1278--1286, 2014.

\bibitem{ldm}
Robin Rombach, Andreas Blattmann, Dominik Lorenz, Patrick Esser, and Björn
  Ommer.
\newblock High-resolution image synthesis with latent diffusion models.
\newblock 2022.

\bibitem{palette}
Chitwan Saharia, William Chan, Huiwen Chang, Chris Lee, Jonathan Ho, Tim
  Salimans, David Fleet, and Mohammad Norouzi.
\newblock Palette: Image-to-image diffusion models.
\newblock In {\em SIGGRAPH}, 2022.

\bibitem{3dp2020}
Meng-Li Shih, Shih-Yang Su, Johannes Kopf, and Jia-Bin Huang.
\newblock 3d photography using context-aware layered depth inpainting.
\newblock In {\em Proc. IEEE Conf. Comput. Vis. Pattern Recognit.}, 2020.

\bibitem{ibr}
Heung-Yeung Shum, Shing-Chow Chan, and Sing~Bing Kang.
\newblock {\em Image-based rendering}.
\newblock Springer Science and Business Media, 2008.

\bibitem{diff_model1}
Jascha Sohl-Dickstein, Eric~A. Weiss, Niru Maheswaranathan, and Surya Ganguli.
\newblock Deep unsupervised learning using nonequilibrium thermodynamicss.
\newblock In {\em Proc. Int. Conf. Mach. Learn.}, 2015.

\bibitem{NeuralIllum}
Shuran Song and Thomas Funkhouser.
\newblock Neural illumination: Lighting prediction for indoor environmentsk.
\newblock In {\em Proc. IEEE Conf. Comput. Vis. Pattern Recognit.}, pages
  6918--6926, 2019.

\bibitem{Im2Pano3D}
Shuran Song, Andy Zeng, Angel~X. Chang, Manolis Savva, Silvio Savarese, and
  Thomas Funkhouser.
\newblock Im2pano3d: Extrapolating 360$^{\circ}$ structure and semantics beyond
  the field of view.
\newblock In {\em Proc. IEEE Conf. Comput. Vis. Pattern Recognit.}, pages
  3847--3856, 2018.

\bibitem{autoint}
Weiping Song, Chence Shi, Zhiping Xiao, Zhijian Duan, Yewen Xu, Ming Zhang, and
  Jian Tang.
\newblock Autoint: Automatic feature interaction learning via self-attentive
  neural networks.
\newblock In {\em Proc. IEEE Conf. Comput. Vis. Pattern Recognit.}, 2021.

\bibitem{Lighthouse}
Pratul~P. Srinivasan, Ben Mildenhall, Matthew Tancik, Jonathan~T. Barron,
  Richard Tucker, and Noah Snavely.
\newblock Lighthouse: Predicting lighting volumes for spatially-coherent
  illumination.
\newblock In {\em Proc. IEEE Conf. Comput. Vis. Pattern Recognit.}, pages
  8080--8089, 2020.

\bibitem{panoSyn_j}
Julius~Surya Sumantri and In~Kyu Park.
\newblock 360 panorama synthesis from a sparse set of images on a low-power
  device.
\newblock {\em IEEE Trans. on Comput. Imaging}, 6:1179--1193, 2020.

\bibitem{panoSyn}
Julius~Surya Sumantri and In~Kyu Park.
\newblock 360 panorama synthesis from a sparse set of images with unknown fov.
\newblock In {\em IEEE Winter Conf. Applications of Comput. Vis.}, pages
  2386--2395, 2020.

\bibitem{lama}
Roman Suvorov, Elizaveta Logacheva, Anton Mashikhin, Anastasia Remizova,
  Arsenii Ashukha, Aleksei Silvestrov, Naejin Kong, Harshith Goka, Kiwoong
  Park, and Victor Lempitsky.
\newblock Resolution-robust large mask inpainting with fourier convolutions.
\newblock In {\em IEEE Winter Conf. Applications of Comput. Vis.}, 2022.

\bibitem{inception_v3}
Christian Szegedy, Vincent Vanhoucke, Sergey Ioffe, Jonathon Shlens, and
  Zbigniew Wojna.
\newblock Rethinking the inception architecture for computer vision.
\newblock In {\em Proc. IEEE Conf. Comput. Vis. Pattern Recognit.}, 2016.

\bibitem{block-nerf}
Matthew Tancik, Vincent Casser, Xinchen Yan, Sabeek Pradhan, Ben Mildenhall,
  Pratul~P. Srinivasan, Jonathan~T. Barron, and Henrik Kretzschmar.
\newblock Block-nerf: Scalable large scene neural view synthesis.
\newblock {\em arXiv:2202.05263}, 2022.

\bibitem{neural_rendering}
Ayush Tewari, Theobalt Christian, Dan~B Goldman, Eli Shechtman, Gordon
  Wetzstein, Jason Saragih, Jun-Yan Zhu, Justus Thies, Kalyan Sunkavalli,
  Maneesh Agrawala, Matthias Niessner, Michael Zollhofer, Ohad Fried,
  Ricardo~Martin Brualla, Rohit~Kumar Pandey, Sean Fanello, Stephen Lombardi,
  Tomas Simon, and Vincent Sitzmann.
\newblock State of the art on neural rendering.
\newblock {\em Computer Graphics Forum}, 2020.

\bibitem{grf2021}
Alex Trevithick and Bo Yang.
\newblock Grf: Learning a general radiance field for 3d scene representation
  and rendering.
\newblock In {\em International Conference on Computer Vision}, 2021.

\bibitem{mega-nerf}
Haithem Turki, Deva Ramanan, and Mahadev Satyanarayanan.
\newblock Mega-nerf: Scalable construction of large-scale nerfs for virtual
  fly-throughs.
\newblock In {\em Proc. IEEE Conf. Comput. Vis. Pattern Recognit.}, 2022.

\bibitem{ibr-net}
Qianqian Wang, Zhicheng Wang, Kyle Genova, Pratul Srinivasan, Howard Zhou,
  Jonathan~T. Barron, Ricardo Martin-Brualla, Noah Snavely, and Thomas
  Funkhouser.
\newblock Ibrnet: Learning multi-view image-based rendering.
\newblock In {\em Proc. IEEE Conf. Comput. Vis. Pattern Recognit.}, 2021.

\bibitem{ssim}
Zhou Wang, Alan~C. Bovik, Hamid~R. Sheikh, and Eero~P. Simoncelli.
\newblock Image quality assessment: from error visibility to structural
  similarity.
\newblock {\em IEEE Trans. Image Process}, 13(4):600--612, 2004.

\bibitem{nerf--}
Zirui Wang, Shangzhe Wu, Weidi Xie, Min Chen, and Victor~Adrian Prisacariu.
\newblock Nerf--: Neural radiance fields without known camera parameters.
\newblock {\em arXiv:2102.07064}, 2021.

\bibitem{GAN_IC4}
Z. Yan, X. Li, M. Li, W. Zuo, and S. Shan.
\newblock Shift-net: Image inpainting via deep feature rearrangement.
\newblock In {\em Proc. Eur. Conf. Comput. Vis.}, pages 1--17, 2018.

\bibitem{plenoctrees}
Alex Yu, Ruilong Li, Matthew Tancik, Hao Li, Ren Ng, and Angjoo Kanazawa.
\newblock Plenoctrees for real-time rendering of neural radiance fields.
\newblock In {\em International Conference on Computer Vision}, 2021.

\bibitem{pixel-nerf}
Alex Yu, Vickie Ye, Matthew Tancik, and Angjoo Kanazawa.
\newblock Pixelnerf: Neural radiance fields from one or few images.
\newblock In {\em Proc. IEEE Conf. Comput. Vis. Pattern Recognit.}, 2021.

\bibitem{GAN_IC3}
J. Yu, Z. Lin, J. Yang, X. Shen, X. Lu, and T.~S. Huang.
\newblock Generative image inpainting with contextual attention. in proceedings
  of the ieee conference on computer vision and pattern recognition.
\newblock In {\em Proc. IEEE Conf. Comput. Vis. Pattern Recognit.}, pages
  5505--5514, 2018.

\bibitem{GAN_IC6}
J. Yu, Z. Lin, J. Yang, X. Shen, X. Lu, and T.~S. Huang.
\newblock Free-form image inpainting with gated convolution.
\newblock In {\em Proc. IEEE Int. Conf. Comput. Vis.}, pages 4471--4480, 2019.

\bibitem{dii_bat}
Yingchen Yu, Fangneng Zhan, Rongliang Wu, Jianxiong Pan, Kaiwen Cui, Shijian
  Lu, Feiying Ma, Xuansong Xie, and Chunyan Miao.
\newblock Diverse image inpainting with bidirectional and autoregressive
  transformers.
\newblock In {\em Proc. ACM Int. Conf. on Multimedia}, 2021.

\bibitem{nerf-pp}
Kai Zhang, Gernot Riegler, Noah Snavely, and Vladlen Koltun.
\newblock Nerf++: Analyzing and improving neural radiance fields.
\newblock {\em arXiv:2010.07492}, 2020.

\bibitem{lpips}
Richard Zhang, Phillip Isola, Alexei~A. Efros, Eli Shechtman, and Oliver Wang.
\newblock The unreasonable effectiveness of deep features as a perceptual
  metric.
\newblock In {\em Proc. IEEE Conf. Comput. Vis. Pattern Recognit.}, 2018.

\bibitem{pcssc}
Shoulong Zhang, Shuai Li, Aimin Hao, and Hong Qin.
\newblock Point cloud semantic scene completion from rgb-d images.
\newblock In {\em Proc. AAAI Conf. Artif. Intell.}, pages 3385--3393, 2021.

\bibitem{ddcomp}
Yinda Zhang and Thomas Funkhouser.
\newblock Deep depth completion of a single rgb-d image.
\newblock In {\em Proc. IEEE Computer Vision and Pattern Recognition}, 2018.

\bibitem{UCTGAN}
Lei Zhao, Qihang Mo, Sihuan Lin, Zhizhong Wang, Zhiwen Zuo, Haibo Chen, Wei
  Xing, and Dongming Lu.
\newblock Uctgan: Diverse image inpainting based on unsupervised cross-space
  translation.
\newblock In {\em Proc. IEEE Conf. Comput. Vis. Pattern Recognit.}, pages
  5741--5750, 2020.

\bibitem{comodgan}
S. Zhao, J. Cui, Y. Sheng, Y. Dong, X. Liang, E.~I. Chang, and Y. Xu.
\newblock Large scale image completion via co-modulated generative adversarial
  networks.
\newblock In {\em Proc. Int. Conf. Learn. Representations}, 2021.

\bibitem{PIC}
Chuanxia Zheng, Tat-Jen Cham, and Jianfei Cai.
\newblock Pluralistic image completion.
\newblock In {\em Proc. IEEE Conf. Comput. Vis. Pattern Recognit.}, pages
  1438--1447, 2020.

\bibitem{str3d}
Jia Zheng, Junfei Zhang, Jing Li, Rui Tang, Shenghua Gao, and Zihan Zhou.
\newblock Structured3d: A large photo-realistic dataset for structured 3d
  modeling.
\newblock In {\em Europ. Conf. Comput. Vis.}, 2020.

\bibitem{mpi}
Tinghui Zhou, Richard Tucker, John Flynn, Graham Fyffe, and Noah Snavely.
\newblock Stereo magnification: Learning view synthesis using multiplane
  images.
\newblock In {\em ACM Trans. Graph}, 2018.

\bibitem{sparse_fusion}
Zhizhuo Zhou and Shubham Tulsiani.
\newblock Sparsefusion: Distilling view-conditioned diffusion for 3d
  reconstruction.
\newblock In {\em Proc. IEEE Conf. Comput. Vis. Pattern Recognit.}, 2023.

\end{thebibliography}
}

\appendix

\section{Optimization method for complete images selection}
\label{sec:detail_ic}

\subsection{Algorithm}

The optimization problem for selecting a subset of the completed images formulated in \refEq{problem} is solved through  simulated annealing (SA) \cite{sim_ann}. SA does not only proceed with the search in the direction of higher values of the evaluation function, but also adopts a solution with a specific probability even when the evaluation value becomes worse. The probability of selecting a modified solution is controlled by the temperature parameter $T$. In the early stages of the search, the temperature is high and a large area is explored, while at the end of the search, the temperature is low and local search is approached. This allows finding a global suboptimal solution without falling into a local solution.

Using the camera pose of the completes images and the positions of the completed regions as input, the algorithm outputs a subset of the completed images to be used for training NeRF.  \refFig{sa_flow} shows the optimization process flow, which is described below.

\putFigW{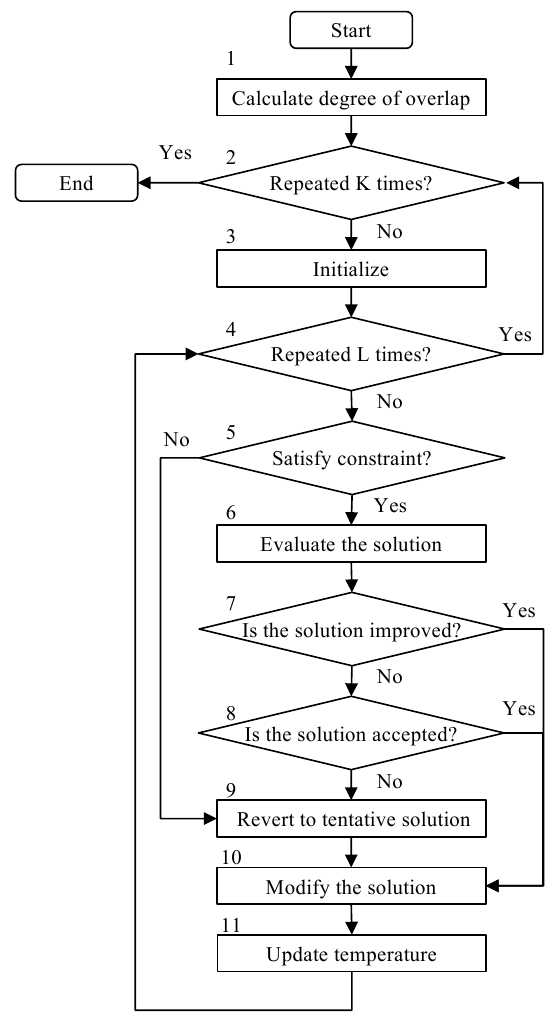}{The optimization process flow for the complete images selection}{70mm}

\textbf{Step 1:} We calculate the degree of overlap between the completed images using the camera pose of the completes images and the positions of the completed regions according to \refEq{p_ijkl}, (\ref{eq:d_ijkl}) and (\ref{eq:r_ij}).

\textbf{Step 2:} The initial solution is changed $K=10$ times and optimization by SA is performed; if it is repeated $K$ times, the process is terminated and the best solution in the search is outputted; otherwise, proceed to the next step.

\textbf{Step 3:} Initialize the SA parameters and solution. The temperature $T$ is set to $1.0 \times 10^3$, and the temperature attenuation factor $\eta$ is set to 0.9995. A randomly selected one completed image is set as the initial solution, and the one that includes no completed images is set as the tentative solution.

\textbf{Step 4:} $L=20,000$ iterations were used to terminate one search. if it is repeated $L$ times, go to Step 2; otherwise, proceed to the next step.

\textbf{Step 5:} Determine if the current solution, a subset of the completed images, has no overlap based on the constraint \refEq{constraint}. If the constraints are satisfied, proceed to the next step; otherwise, go to Step 9.

\textbf{Step 6:} Calculate the value of the evaluation function in \refEq{problem}, i.e., the number of rays in the completed area of the completed images selected as the solution.

\textbf{Step 7:} If the current solution evaluation value is higher than the tentative solution evaluation value, the tentative solution is updated with the current solution and go to Step 10. Otherwise, go to the next step.

\textbf{Step 8:} Accept the current solution with probability $\exp(d/T)$, where $d$ is the evaluation value of the current solution minus the tentative evaluation value. If the current solution is accepted, the tentative solution is updated with the current solution and go to Step 10. Otherwise, go to the next step.

\textbf{Step 9:} Revert the current solution to the tentative solution.

\textbf{Step 10:} Modify the current solution. We randomly selects a new completed image to add to the solution with 10\% probability, randomly deletes a completed image in the current solution with 10\% probability, and replaces completed images in the current solution with the completed images in the neighborhood with 80\% probability. For the replacement of completed images in the neighborhood, the completed images have the indices of $\{1, 2, \cdots,10\} \times \{1, 2, \cdots,10\}$ grid, and the random numbers sampled from the standard normal distribution are added to the indices and rounded to integer values.

\textbf{Step 11:} The temperature $T$ is updated as $T \leftarrow \eta T$, and return to step4.

\subsection{Processing time}

We measured the processing times of the above optimization algorithms on the Tesla V100. Ten trials were performed, and the average processing time was 222.1 sec, standard deviation 148.4 sec, and maximum 488.3 sec. The variation in processing time is due to the fact that the computational complexity is on the order of squared for the number of completed images selected as the solution.

\subsection{Selected completed images}

Figure \ref{fig:comp_proc} shows an example of the content and location of the completed images selected to train NeRF. Notably, although not indicated in this figure, 100 non-completed reprojection images were also utilized to train NeRF, as in OmniNeRF \cite{omninerf}. From this figure, it can be seen that the completed images are selected to be widely distributed in space to cover the occlusion region. Many images are selected near the input image position (0, 0), because the completed region is smaller near the input image, resulting in less overlap.

\putFigWW{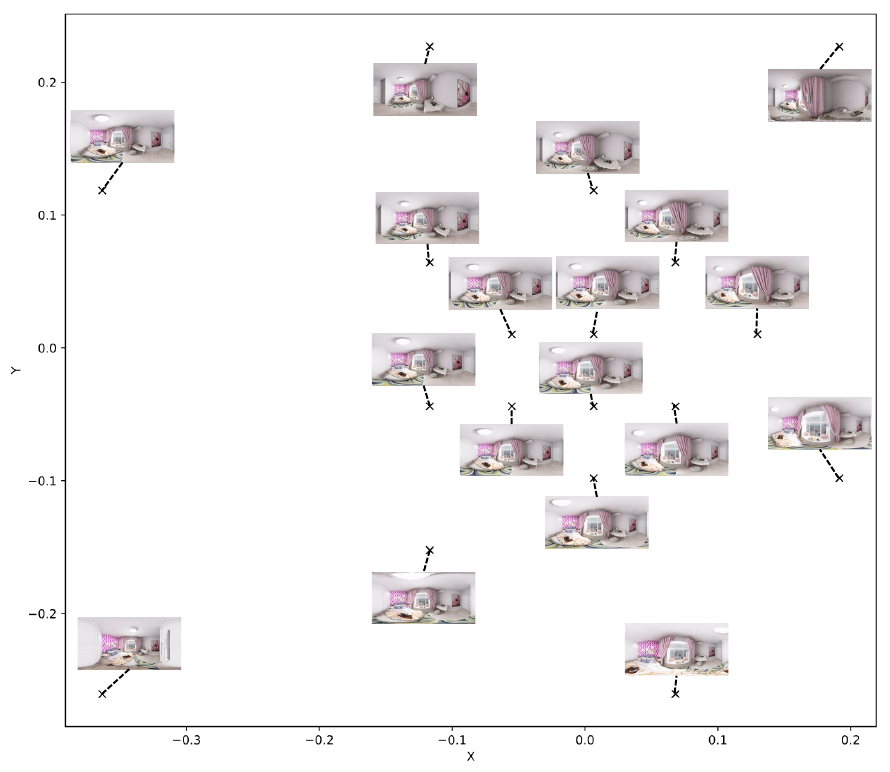}{Example of the content and location of the completed images selected to train NeRF. The $X$- and $Y$-axis lie on a plane that is orthogonal to gravity axis $Z$, and the maximum depth in the input image is scaled to 1.0. The each completed image exists on a grid equally divided by 10 points in the x-axis interval [-0.363, 0.192] and y-axis interval [-0.261, 0.227], respectively.}{170mm}

\section{Details of image completion}
\label{sec:detail_ic}

\subsection{Settings}

We trained the network using the training data in Section \ref{sec:dataset} and 3,200 masks of missing regions. 32 scenes are selected from the training data in both Structured3D and Matterport3D dataset and each 360-degree RGB-D image is reprojected onto the 10 $\times$ 10 grid shown in Section \ref{sec:impl} to generate masks of missing regions. As a data augmentation, the training image and mask were randomly rotated around the gravity axis, which corresponds to a horizontal cyclic shift on the equirectangular image. We utilized OmniDreamer \cite{omni_dreamer} as the image completion network and default settings in \cite{omni_dreamer}  are used for training and model hyper-parameters.

\subsection{Completed images}

\putFigWW{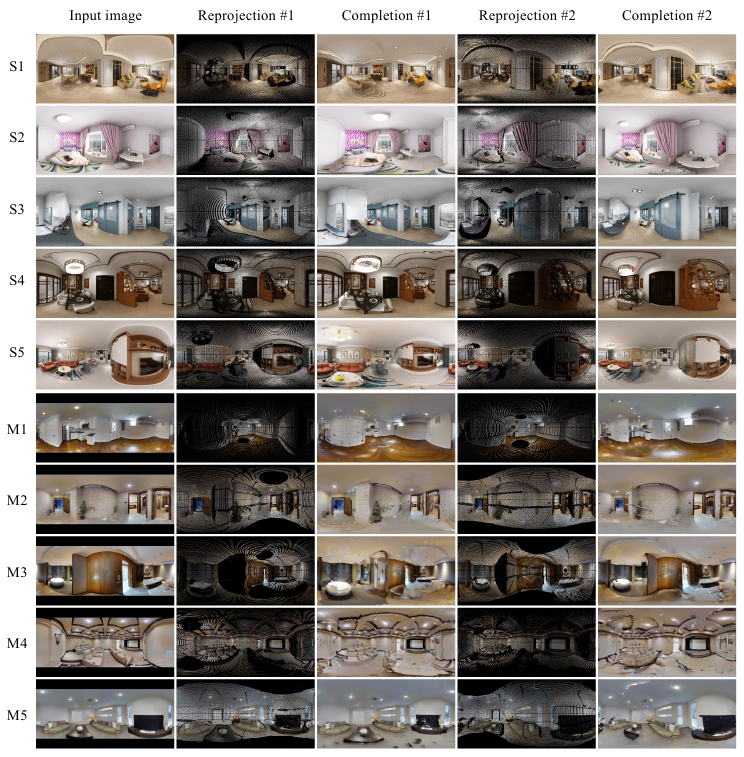}{Examples of the completed images. Black pixels on the reprojected images are missing pixels due to occlusion or resolution change.}{174mm}

Examples of images in which the missing pixels in the reprojected images are completed by OmniDreamer is shown in \refFig{sample_comp}.

\section{Details of NeRF}

We trained the NeRF model using the Adam optimizer \cite{adam} while exponentially decreasing the learning rate from $5.0\times10^{-4}$ to $5.0\times10^{-5}$. The NeRF model was trained with 200,000 iterations for each experiment with a batch size of 1,400. We set the number of sampling $N_c$ = 64 and $N_f$ = 128 for the coarse and refined networks, respectively. The network structures of NeRF were identical to those of Instant-NGP \cite{hash-nerf}. Learning one scene took approximately 4 hours on an NVIDIA Tesla V100 GPU. 

\section{NLL for synthesized images}

\putFigWW{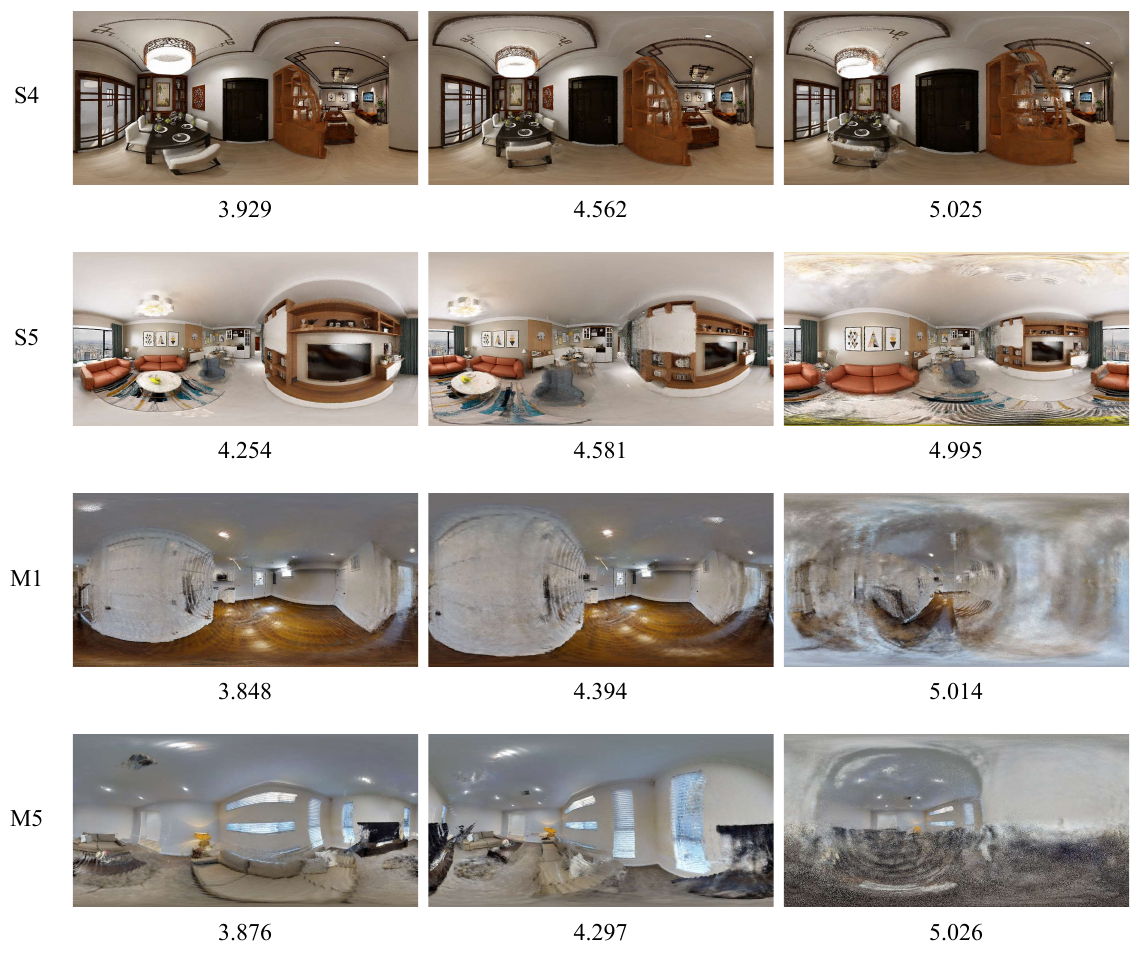}{Examples of images and NLL values. Note that the absolute value of NLL has no meaning across data sets because the likelihood models are different for Structured3D dataset and Matterport3D dataset.}{150mm}

\subsection{Definition of NLL}

We employ the negative log likelihood (NLL) for quantitative evaluation of the image quality of novel views. The $D=2048$ dimensional values of the last pooling layer of inception-v3 \cite{inception_v3} are used as features, their distribution is modeled with a normal distribution. The mean $\mu_f \in \mathbb{R}^{D}$ and covariance matrix $\Sigma_f \in \mathbb{R}^{D \times D}$ of the distribution were obtained from the features of 20,000 randomly cropped perspective images from the test data by maximum likelihood estimation. Using the features $\{x_i \in \mathbb{R}^{D}\}_{i=1}^{M}$ of randomly synthesizing $M = 2,000$ perspective projection images from the learned 3D scene, and the NLL is calculated by the following formula:
\begin{equation}
\label{eq:nllf}
{\rm NLL} = \frac{1}{MD} \sum_{i=1}^{M} (x_i - \mu_f) ^T \Sigma_f^{-1} (x_i - \mu_f) 
\end{equation}

\subsection{Validity of NLL}

To confirm the validity of this metric, examples of images and NLL values are shown in \refFig{nll}. In this figure, NLL is calculated using 512 perspective images  randomly cropped from a single 360-degrees image. The smaller the value of NLL, the better the perceived image quality appears.

\section{Additional Results}
\label{sec:add_results}

\subsection{Synthesized novel views}

\putFigWW{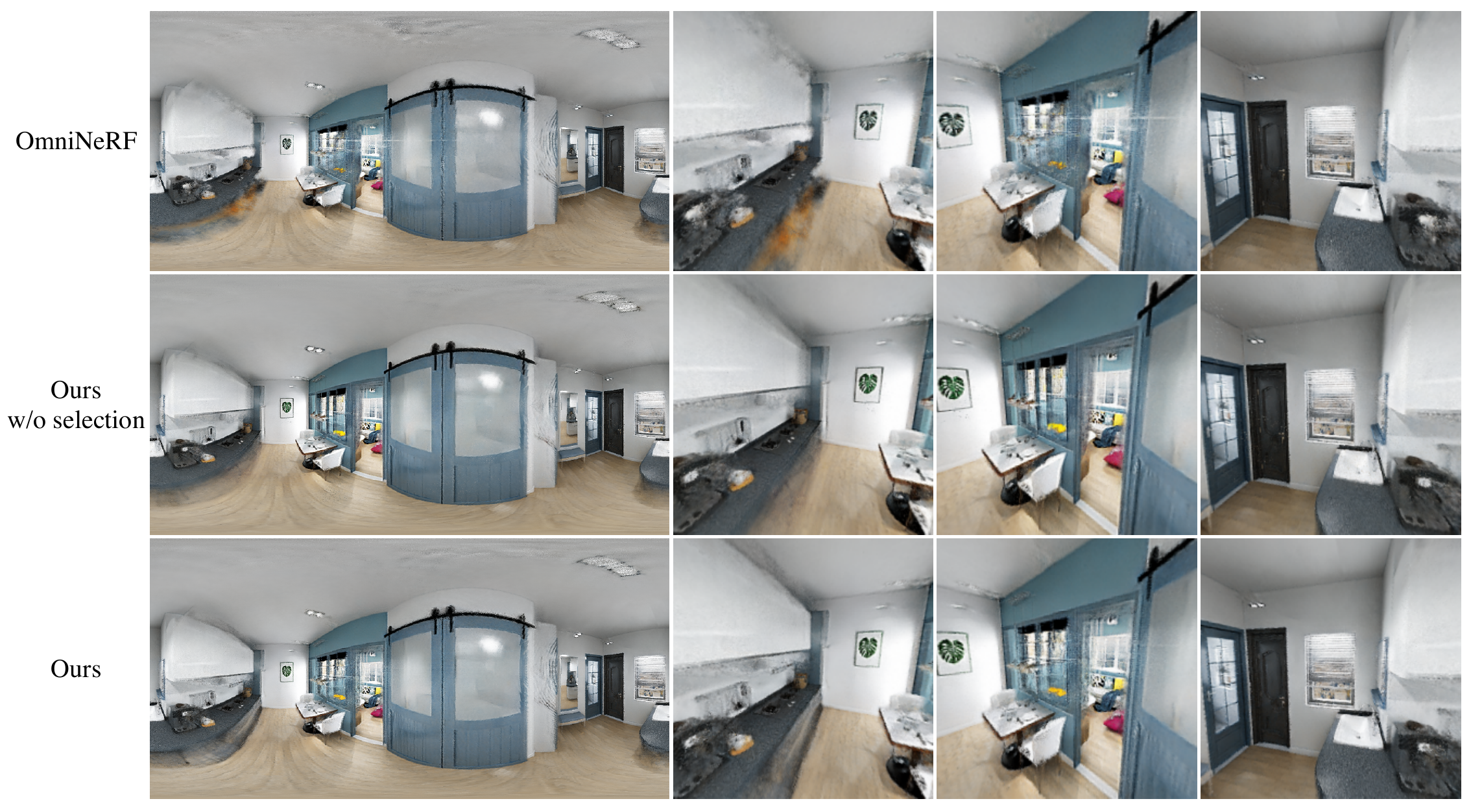}{Novel views synthesized in equirectangular and perspective projections with different camera positions for scenes S3.}{160mm}

\putFigWW{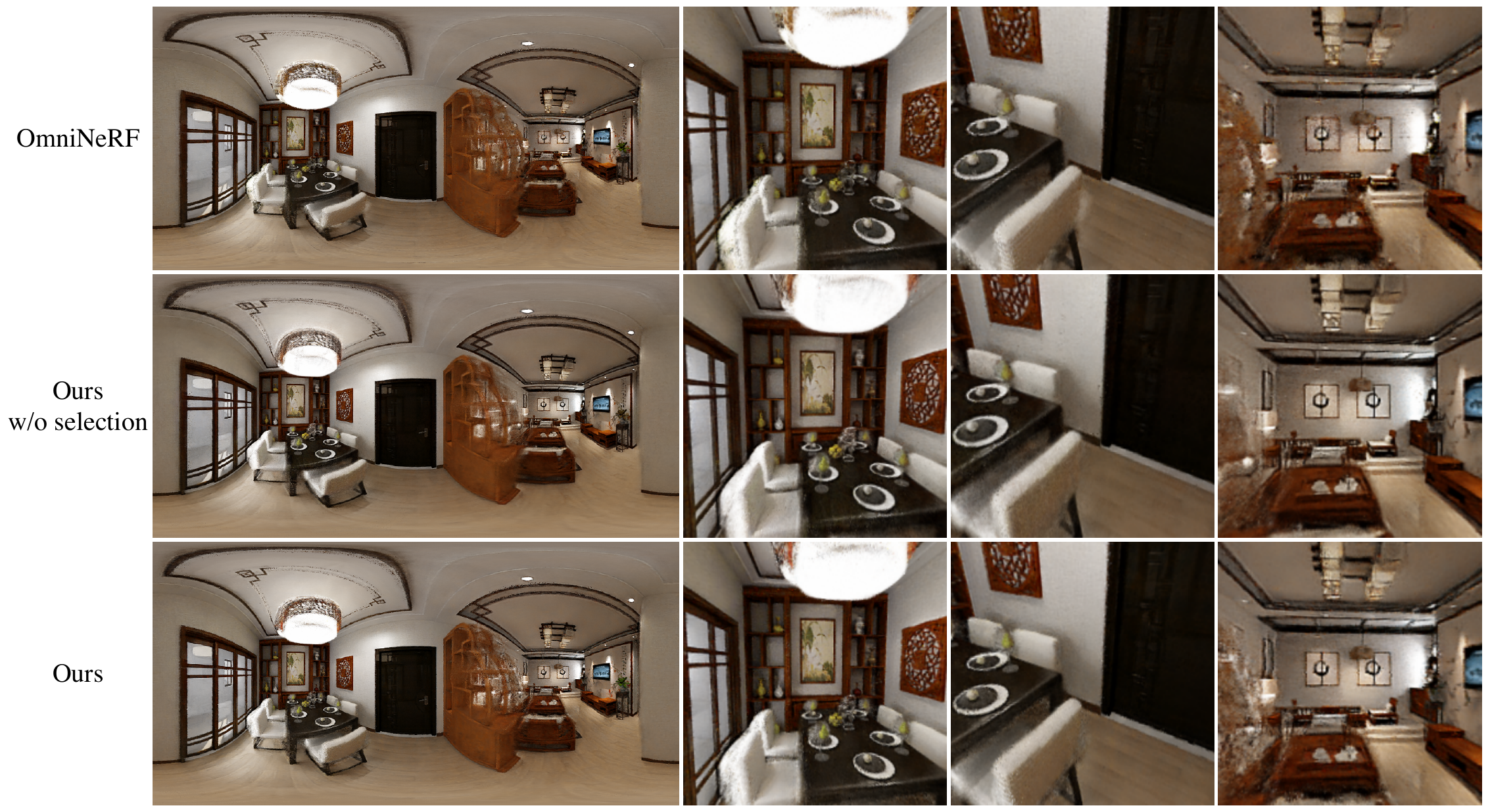}{Novel views synthesized in equirectangular and perspective projections with different camera positions for scenes S4.}{160mm}

\putFigWW{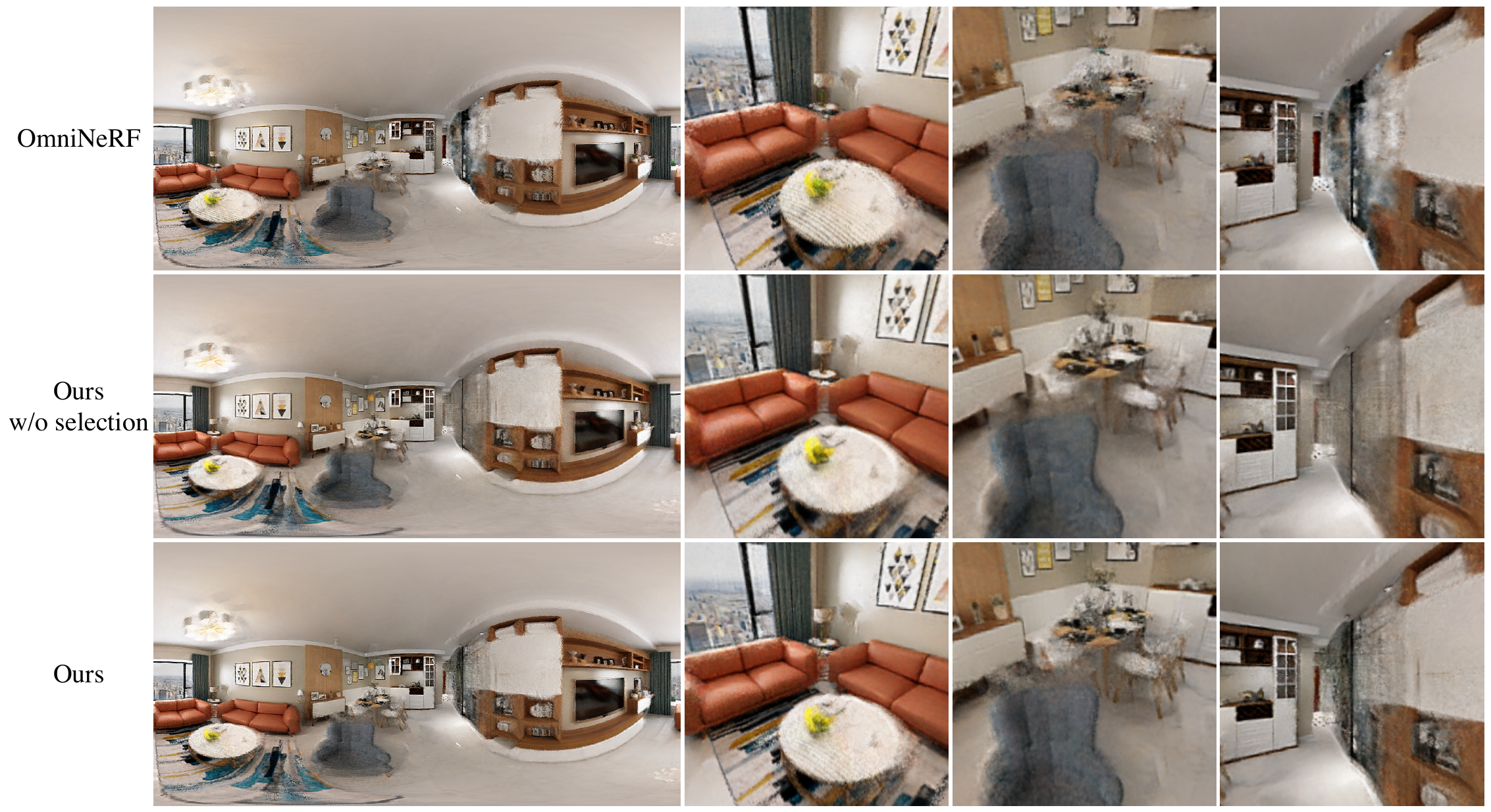}{Novel views synthesized in equirectangular and perspective projections with different camera positions for scenes S5.}{160mm}

\putFigWW{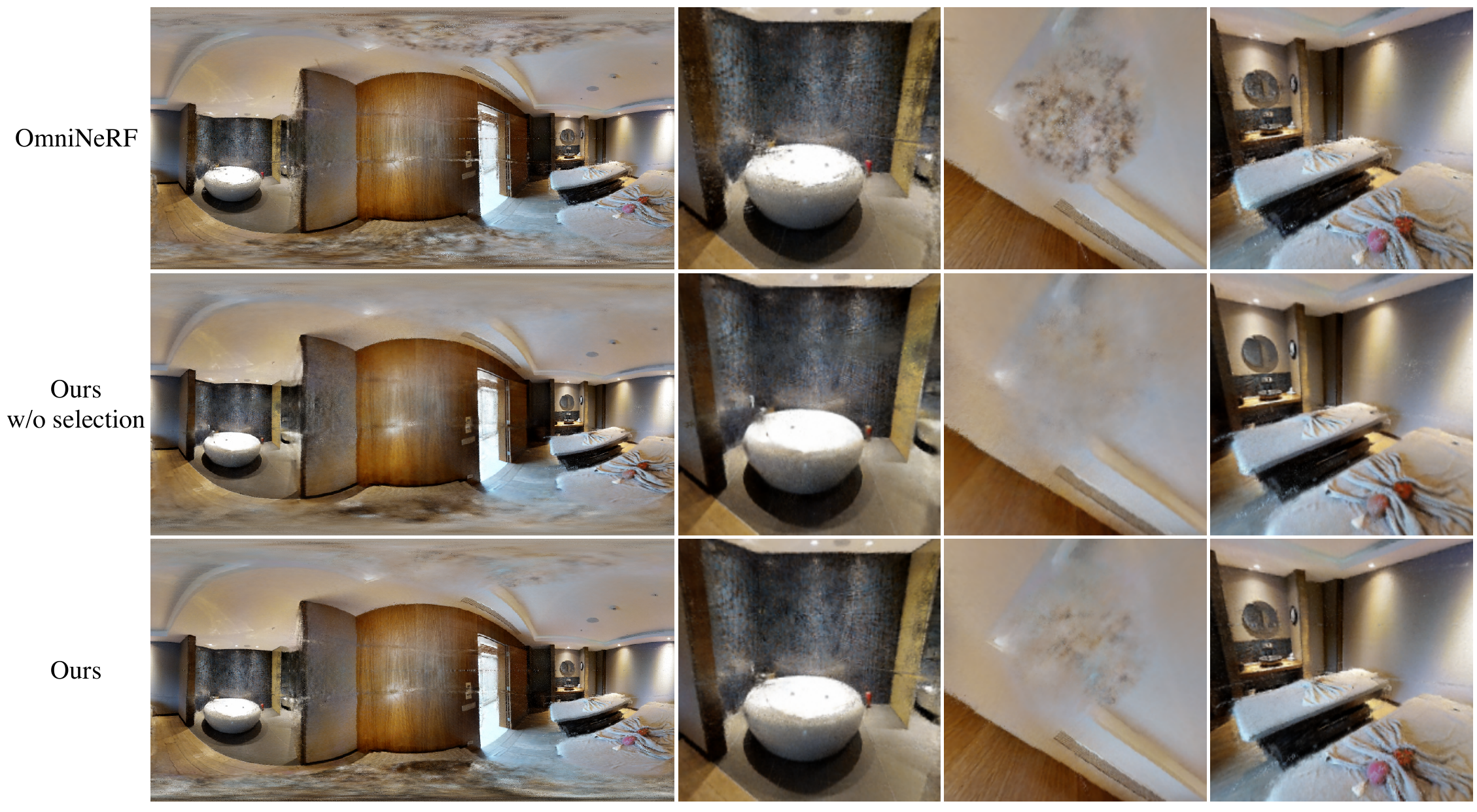}{Novel views synthesized in equirectangular and perspective projections with different camera positions for scenes M3.}{160mm}

\putFigWW{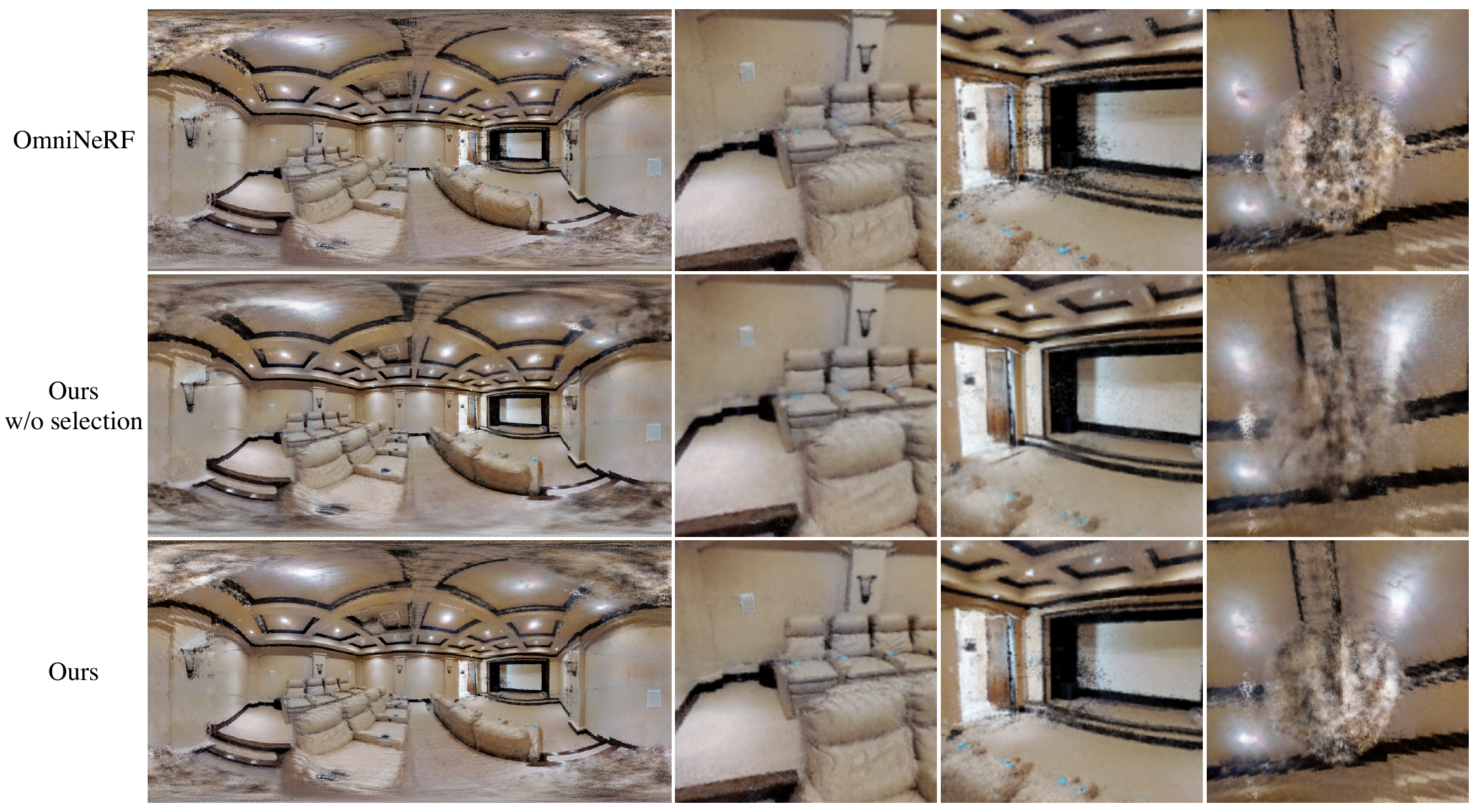}{Novel views synthesized in equirectangular and perspective projections with different camera positions for scenes M4.}{160mm}

\putFigWW{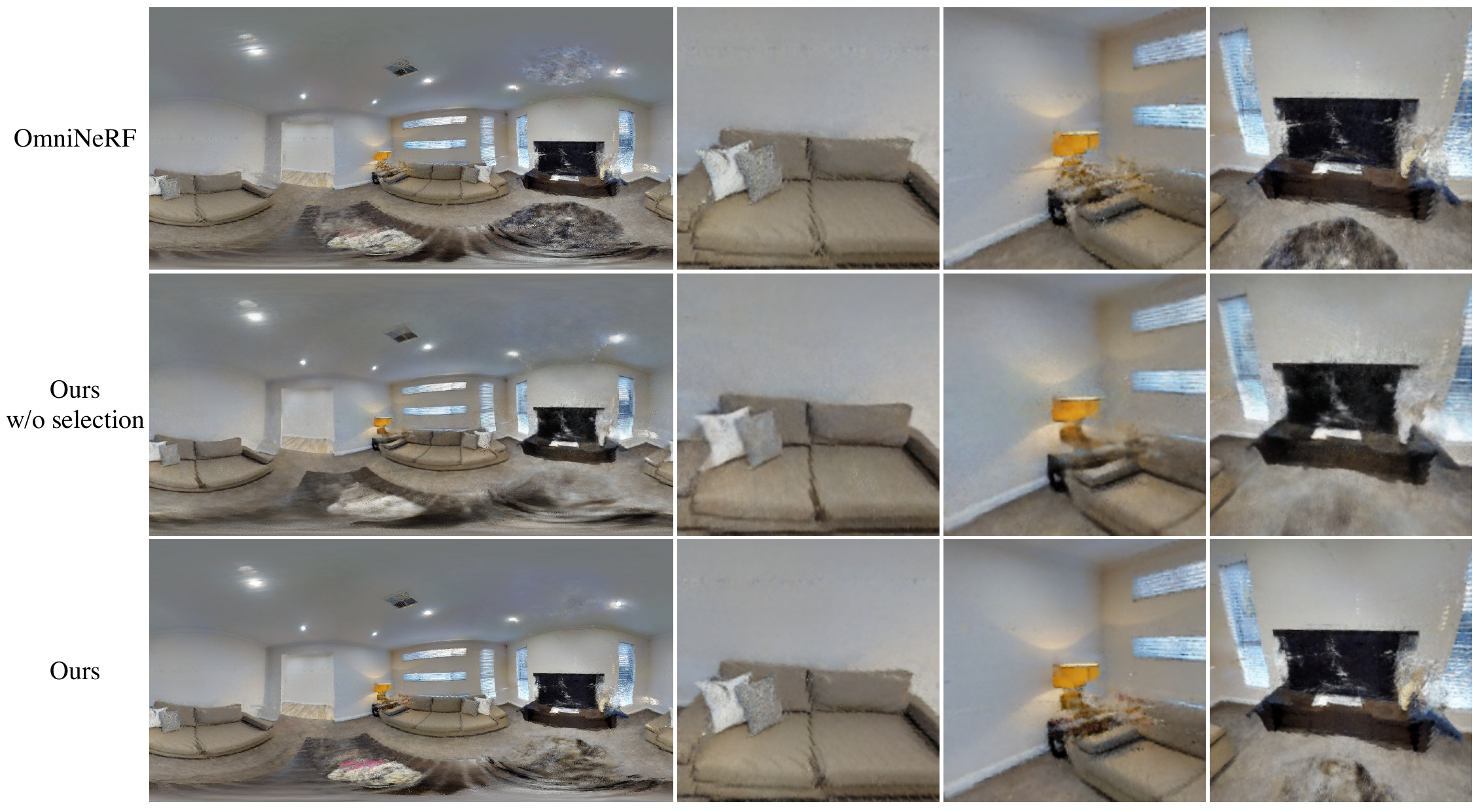}{Novel views synthesized in equirectangular and persptective projections with different camera positions for scenes M5.}{160mm}

The results of novel view synthesis using OmniNeRF \cite{omninerf} and the proposed method (with and without the completed images selection) in scenes S3, S4, S5, M3, M4, and M5 are illustrated in \refFig{sample_s3}, \ref{fig:sample_s4}, \ref{fig:sample_s5}, \ref{fig:sample_m3}, \ref{fig:sample_m4} and \ref{fig:sample_m5}. The figures show synthesized novel views in equirectangular and perspective projections with different camera positions.
\subsection{Rendering from a free moving camera}

Examples of rendering from a freely moving camera are shown in \refFig{free_moving}. In the figure, novel views are rendered in perspective projection with a horizontal field of view of 90 degrees.

\putFigWW{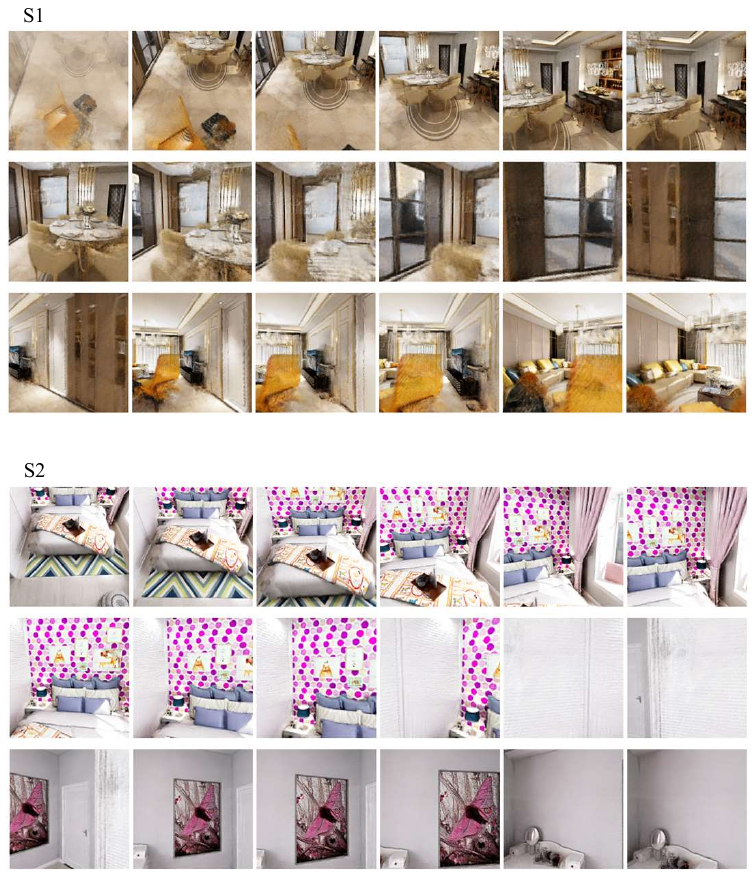}{Examples of rendering from a freely moving camera. The novel views are rendered in perspective projection with a horizontal field of view of 90°.}{160mm}

\end{document}